\documentclass[
    a4paper,
    man,
    floatsintext
]{glossaPX2}

\usepackage[T1]{fontenc}
\usepackage[american]{babel}
\usepackage[style=apa,backend=biber,sorting=nyt,natbib=true]{biblatex}
\NewBibliographyString{unpublished}
\DefineBibliographyStrings{english}{unpublished = {Unpublished}}
\DefineBibliographyStrings{american}{unpublished = {Unpublished}}
\usepackage[font={footnotesize,it}]{caption}
\usepackage{csquotes}
\usepackage{booktabs}
\usepackage{tabularx}
\usepackage{linguex}
\usepackage{cgloss}

\usepackage{amsmath,amssymb,amsfonts,amsthm,mathrsfs}
\makeatletter
\tagsleft@false
\makeatother
\usepackage{graphicx}
\graphicspath{{figures/}}
\usepackage{ragged2e}
\usepackage{hyperref}
\usepackage{enumitem}
\usepackage[title]{appendix}
\usepackage{multirow}
\usepackage{makecell}
\usepackage{array}
\usepackage{adjustbox}
\usepackage{colortbl}
\usepackage{xcolor}
\usepackage{algorithm}
\usepackage{algpseudocode}
\usepackage{listings}
\usepackage{cleveref}
\usepackage{placeins}
\usepackage{float}
\usepackage{subcaption}
\usepackage{rotating}
\usepackage{url}
\usepackage{microtype}

\newcolumntype{L}{@{\extracolsep{\fill}}l}
\newcolumntype{C}{@{\extracolsep{\fill}}c}
\newcolumntype{R}{@{\extracolsep{\fill}}r}
\title[Fusion-Aware Direct 3D Gaussian Generation with Structured Patch Latent Flows]{Fusion-Aware Direct 3D Gaussian Generation with Structured Patch Latent Flows}
\author[Wang et al.]
{\spauthor{Yizhao Wang\\
  \institute{School of Computer Science and Technology, Henan Institute of Science and Technology}\\
  \small{cswyz@stu.hist.edu.cn, ORCID: 0009-0003-7056-7264}
  }
  \AND
\spauthor{Jingbo Wang\\
  \institute{School of Computer Science and Technology, Henan Institute of Science and Technology}\\
  \small{jingbowang@hist.edu.cn}
  }
  \AND
\spauthor{Guantao Zhang\\
  \institute{School of Computer Science and Technology, Henan Institute of Science and Technology}\\
  \small{guantaozhang@hist.edu.cn}
  }}

\begin{document}
\maketitle

\begin{abstract}
Class-guided 3D object generation is important for intelligent content creation, virtual environments, and digital asset design. Although 3D Gaussian Splatting (3DGS) offers an explicit and render-efficient representation, directly generating 3D Gaussian objects is difficult because Gaussian primitives are unordered, variable-sized, locally dense, and highly sensitive to rendering. Existing 3DGS generation methods usually depend on multi-view synthesis, reconstruction, or lifted 2D priors, fusing information mainly from observed views rather than modeling the intrinsic structural distribution of 3D Gaussian objects.

This paper proposes a fusion-aware hierarchical Gaussian patch representation for direct class-guided 3DGS generation with rectified flow. Irregular Gaussian sets are decomposed into canonical local patches and encoded as structured tokens. The resulting hierarchical latent space fuses global class semantics, patch-level geometry and appearance, spatial correspondence, and rendering-sensitive cues. On this basis, we design a structure-aware rectified flow model with patch-position conditioning, global-local coupled velocity prediction, and density-aware velocity weighting, enabling direct latent generation of class-conditioned 3DGS objects within seconds. A render-feedback fusion strategy further aligns latent flow learning with decoded multi-view rendering quality.

Experiments show that the proposed method generates 3D Gaussian objects with more coherent geometry, sharper local details, and better multi-view consistency than baseline latent generative models. Ablation studies confirm the contributions of hierarchical information fusion, global-local coupling, density-aware supervision, and render-feedback learning while preserving practical sampling efficiency overall.
\end{abstract}


\section{Introduction}

3D object generation has become a fundamental technique for intelligent content creation, virtual environments, digital asset design, and embodied perception. Early 3D generative models mainly relied on voxels, point clouds, and implicit fields, including adversarial shape generation, point cloud autoencoding, flow-based point generation, and diffusion-based point cloud synthesis. These methods have substantially advanced class-conditioned 3D object modeling, yet they still face trade-offs among representation compactness, geometric fidelity, appearance quality, and computational efficiency. In particular, point-based representations are flexible but weak in appearance modeling, voxel-based methods are structured but memory-intensive, and implicit neural fields provide continuous geometry and radiance at the cost of expensive optimization and rendering.

Neural radiance fields have further improved differentiable 3D representation and novel-view synthesis by encoding geometry and appearance in continuous neural fields~\citep{mildenhall2020nerf}. Combined with pretrained image diffusion models~\citep{ho2020denoising,rombach2022high}, recent text- or image-guided 3D generation methods have achieved impressive progress by optimizing 3D representations with lifted 2D priors~\citep{poole2023dreamfusion,shi2024mvdream}. Although effective for open-vocabulary 3D asset creation, these approaches often depend on per-instance optimization, multi-view synthesis, or image-conditioned reconstruction, and therefore provide limited control over the intrinsic distribution of 3D objects. This limitation becomes more pronounced in class-guided generation, where the objective is to directly learn stable intra-class 3D object distributions from limited 3D data rather than to lift abundant 2D priors into 3D space.

Recently, 3D Gaussian Splatting (3DGS) has emerged as an explicit and render-efficient 3D representation with high visual fidelity and real-time rendering capability~\citep{kerbl2023gaussian,li2026mvgsplatting}. By representing a scene or object as a set of anisotropic Gaussian primitives with learnable position, covariance, opacity, and appearance attributes, 3DGS offers an attractive representation space for efficient 3D content generation. Existing 3DGS-based generation methods have mainly followed two routes: optimization-based text-to-3D generation with Gaussian primitives~\citep{chen2024text3d,yi2024gaussiandreamer,tang2024dreamgaussian}, and structured Gaussian modeling, multi-view fusion, or reconstruction pipelines~\citep{zhang2024gaussiancube,ju2025directtrigs,li2026mvgsplatting,hao2024coarse,liu2025maenet}. However, most existing methods still rely on multi-view synthesis, reconstruction, or lifted 2D supervision. As a result, they focus more on recovering or refining Gaussian objects from image-space cues than on directly learning the intrinsic class-conditioned distribution of 3D Gaussian objects themselves. This gap is non-trivial because a 3DGS object is an unordered, variable-sized, locally dense, and rendering-sensitive primitive set, in which different Gaussians contribute unequally to the final rendered appearance.

In parallel, latent generative modeling has become an effective paradigm for high-dimensional data generation by first compressing data into latent spaces and then learning their distributions using diffusion, flow, or transformer-based models~\citep{rombach2022high,lipman2023flowmatching,liu2023flowstraight}. Among them, rectified flow is particularly appealing because it learns a simple velocity field that transports samples from a prior distribution to the data distribution along nearly straight paths~\citep{liu2023flowstraight,lipman2023flowmatching}. Its efficient training and sampling properties have recently motivated its use in 3D generation, including multi-view 3D Gaussian synthesis~\citep{go2025splatflow}. Nevertheless, directly applying rectified flow to 3DGS object generation remains difficult. A naive global latent code may discard local geometry and appearance details, while structure-agnostic token modeling cannot explicitly preserve spatial correspondence, patch-level semantics, or the density-dependent rendering importance of Gaussian primitives.

To address these challenges, this paper proposes a hierarchical Gaussian patch information representation for class-guided 3D object generation with rectified flow. Different from multi-view synthesis or reconstruction-based 3DGS pipelines, our goal is direct class-conditioned generative modeling in the latent space of structured 3D Gaussian objects. Specifically, we first decompose irregular Gaussian primitives into local patches and encode each patch in a canonical coordinate system, producing structured Gaussian patch tokens that preserve local geometry, appearance, and rendering-relevant attributes. A hierarchical latent space is then built to jointly capture global class semantics and local geometric-appearance details. Based on this representation, we design a class-guided structure-aware rectified flow model with patch-position conditioning, global-local coupled velocity prediction, and density-aware velocity weighting, enabling the learned flow to better respect object layout and rendering contribution. Furthermore, we introduce a render-feedback learning strategy to align latent flow optimization with decoded multi-view rendering quality, so that the generated latents correspond to coherent, detailed, and renderable 3D Gaussian objects.

The main contributions of this paper are summarized as follows:
\begin{itemize}
    \item We propose a hierarchical Gaussian patch information representation that transforms irregular 3D Gaussian sets into structured global-local latent tokens, enabling direct class-guided generative modeling of 3DGS objects.
    \item We develop a class-guided structure-aware rectified flow model with patch-position conditioning, global-local coupled velocity prediction, and density-aware velocity weighting to better model the intrinsic distribution of 3D Gaussian objects.
    \item We introduce a render-feedback learning strategy that connects latent flow optimization with decoded multi-view rendering quality, improving the geometric coherence, local details, and view consistency of generated 3DGS objects.
\end{itemize}

\section{Related Work}

\subsection{3D Object Generation}

Learning-based 3D object generation has been extensively studied across different 3D representations, including voxels, point clouds, meshes, implicit fields, neural radiance fields, and hybrid explicit-implicit representations. Early methods adopted voxel grids because of their regular structure and compatibility with convolutional neural networks. Although voxel-based representations provide a convenient structured domain for neural networks, their cubic memory cost restricts resolution and makes it difficult to model fine geometric details.

Point-cloud-based methods provide a more compact and flexible alternative for 3D shape modeling. They support flexible geometric modeling and have been widely studied in 3D perception and representation learning~\citep{sohail2025advancing,fernandes2021pointcloud}. However, point clouds and surface-like primitive sets mainly describe geometry and usually lack explicit appearance, opacity, and rendering-related attributes, which limits their direct use for high-quality view synthesis.

Implicit neural representations address the resolution limitations of explicit grids by modeling 3D shapes as continuous functions. Neural radiance fields further make differentiable view synthesis possible by coupling density and appearance in a continuous radiance representation~\citep{mildenhall2020nerf}. These methods improve geometric fidelity and usability, but they typically require carefully designed surface extraction, dense sampling, or expensive differentiable rendering pipelines.

With the success of pretrained image diffusion models~\citep{ho2020denoising,rombach2022high}, recent 3D generation methods exploit 2D generative priors for text- or image-guided 3D content creation. Representative methods such as DreamFusion and MVDream demonstrate impressive results by optimizing 3D representations using score distillation or multi-view diffusion priors~\citep{poole2023dreamfusion,shi2024mvdream}. Information Fusion studies further show that multi-view, image-language, and image-point-cloud fusion are effective for 3D shape understanding, 3D object detection, point-cloud completion, and scene reconstruction~\citep{ning2024dilf,hao2024coarse,liu2025maenet,wu2026fusionmamba,zhou2025waterhenerf}. Despite these advances, many of these methods rely on lifted 2D priors, sparse-view reconstruction, or per-instance optimization. In contrast, this work focuses on class-guided 3D Gaussian object generation and studies how the intrinsic distribution of 3DGS objects can be modeled through structured latent information representation.

\subsection{3D Gaussian Splatting for Generative Modeling}

3D Gaussian Splatting (3DGS) has emerged as an explicit and render-efficient representation for real-time radiance field rendering~\citep{kerbl2023gaussian}. A 3DGS scene or object is represented as a set of anisotropic Gaussian primitives with learnable position, covariance, opacity, and appearance attributes. Compared with implicit radiance fields, 3DGS provides fast rasterization, explicit primitive manipulation, and high rendering quality, making it attractive for 3D generation and reconstruction. Recent Information Fusion work has begun to connect Gaussian rendering with multi-view guided densification, while related fusion studies emphasize complementary image, language, and point-cloud cues for 3D perception and reconstruction~\citep{li2026mvgsplatting,ning2024dilf,hao2024coarse,sohail2025advancing,liu2025maenet}. However, a 3DGS object is also an unordered, variable-sized, locally dense, and rendering-sensitive primitive set. These properties make direct distribution modeling more difficult than modeling images, voxels, or fixed-grid features.

Several methods have introduced 3DGS into optimization-based text-to-3D pipelines. Text-to-3D using Gaussian Splatting, GaussianDreamer, and DreamGaussian use Gaussian primitives to improve rendering efficiency and accelerate text-guided 3D content creation~\citep{chen2024text3d,yi2024gaussiandreamer,tang2024dreamgaussian}. LucidDreamer further improves the quality of score-distillation-based generation through interval score matching and adopts 3DGS as the optimized 3D representation~\citep{liang2024luciddreamer}. Differentiable rendering and image-language fusion have also been explored for zero-shot 3D shape understanding~\citep{ning2024dilf}, suggesting that semantic conditions can be aligned with 3D representations through fused visual-language cues. Nevertheless, current Gaussian generation methods are still largely driven by 2D diffusion priors, multi-view synthesis, or per-instance optimization. Therefore, the generated Gaussian objects are not learned from an explicit model of the intrinsic class-conditioned 3DGS distribution.

Another line of research develops feed-forward Gaussian reconstruction models. Splatter Image maps image pixels to 3D Gaussian primitives for ultra-fast single-view reconstruction~\citep{szymanowicz2024splatter}, while pixelSplat reconstructs Gaussian radiance fields from image pairs and introduces differentiable sampling of Gaussian means~\citep{charatan2024pixelsplat}. MVSplat improves sparse-view Gaussian reconstruction by incorporating cost-volume geometry cues from multi-view stereo~\citep{chen2024mvsplat}. GS-LRM, LGM, and GRM scale this idea with large reconstruction models that predict high-quality Gaussian primitives from sparse views or generated multi-view images~\citep{zhang2024gslrm,tang2024lgm,xu2024grm}. Information Fusion studies on multi-view Gaussian densification, neural scene reconstruction, image-point-cloud fusion, and point-cloud completion further highlight the importance of cross-view and cross-modal information integration for recovering reliable 3D structure~\citep{li2026mvgsplatting,zhou2025waterhenerf,hao2024coarse,liu2025maenet,wu2026fusionmamba}. These methods are highly effective for reconstruction-driven 3D generation and sparse-view novel view synthesis. However, their Gaussian primitives are usually tied to input views, image tokens, or pixel-aligned prediction, rather than being generated from a compact intrinsic latent distribution of 3D Gaussian objects.

Recent studies have begun to explore structured Gaussian representations for generative modeling. TriplaneGaussian and DirectTriGS introduce triplane-based Gaussian representations to combine the regularity of triplane features with the rendering efficiency of Gaussian Splatting~\citep{zou2024triplanegaussian,ju2025directtrigs}. GaussianCube converts unstructured Gaussian primitives into a structured voxel-like Gaussian representation, making Gaussian fields more compatible with diffusion-based generation~\citep{zhang2024gaussiancube}. Related Information Fusion work on 3D point-cloud understanding, point-cloud language evaluation, and multi-scale completion also suggests that explicit spatial organization and multimodal semantic alignment are important for robust 3D representation learning~\citep{sohail2025advancing,mariani2026glue3d,fernandes2021pointcloud,wu2026fusionmamba}. These methods show that imposing structure on 3D representations is important for generative modeling. Different from grid-based or triplane-based regularization, our method organizes 3DGS objects into local Gaussian patch tokens encoded in canonical coordinates. This design preserves local Gaussian geometry, appearance, and rendering contribution while avoiding the need to force all primitives into a dense grid or image-aligned feature plane. The proposed hierarchical global-local latent space is therefore specifically designed for class-guided generation of 3D Gaussian objects.

\subsection{Rectified Flow and Structured Latent Generation}

Diffusion models have become a dominant paradigm for image and 3D generation because of their stable training and strong distribution modeling ability~\citep{ho2020denoising,rombach2022high}. However, their iterative denoising process can be computationally expensive, especially when combined with differentiable rendering, multi-view supervision, or per-instance 3D optimization. Flow matching and rectified flow provide an alternative continuous-time generative formulation by learning a velocity field that transports samples from a simple prior distribution to the target data distribution~\citep{lipman2023flowmatching,liu2023flowstraight}. Compared with diffusion-based generation, rectified flow uses a simple regression objective and supports efficient sampling along nearly straight transport trajectories.

Recent advances further improve the theoretical and practical foundations of flow-based generative modeling. Conditional flow matching extends flow matching with simulation-free conditional objectives and optimal-transport-inspired probability paths~\citep{tong2024improving}. Simulation-free score and flow matching has also been developed for learning stochastic dynamics between source and target distributions~\citep{tong2024simulation}. Riemannian flow matching generalizes flow matching to non-Euclidean geometries, showing that flow-based generation can be adapted to structured spaces beyond standard Euclidean domains~\citep{chen2024rfm}. These works indicate that flow models are well suited for learning structured transformations between distributions, which is particularly relevant for 3D data whose latent representations often contain spatial, geometric, and topological structures.

In 3D generation, flow-based models have started to appear in structured latent or multi-view generation pipelines. SplatFlow combines a multi-view rectified flow model with a Gaussian Splatting decoder for 3DGS synthesis~\citep{go2025splatflow}. Its flow model operates over multi-view image, depth, and camera latents, which are then decoded into pixel-aligned Gaussian primitives. This demonstrates the potential of rectified flow for Gaussian-based 3D synthesis, but it still relies on multi-view intermediate representations. In contrast, our method directly learns class-conditioned generative dynamics in a hierarchical Gaussian patch latent space. By incorporating patch-position conditioning, global-local coupled velocity prediction, and density-aware velocity weighting, the proposed structure-aware rectified flow explicitly accounts for spatial correspondence, object-level semantics, local Gaussian details, and rendering importance. This makes the flow model more consistent with the intrinsic structure of 3D Gaussian objects rather than treating the latent tokens as an unordered or structure-agnostic sequence.

\section{Method}
\subsection{Problem Formulation and Overview}
Let $\mathcal{D}=\{(\mathcal{G}^{(n)},c^{(n)})\}_{n=1}^{N_d}$ denote the training set, where $\mathcal{G}^{(n)}$ is a 3D Gaussian object and $c^{(n)} \in \{1,\dots,C\}$ is its class label. The goal of class-guided 3D object generation is to learn a conditional distribution $p(\mathcal{G}\mid c)$, so that a realistic 3D Gaussian object can be generated from a class condition $c$ and a random prior sample. Different from multi-view synthesis or reconstruction-based pipelines, our objective is to directly model the distribution of 3D Gaussian objects in a structured latent space rather than recovering them from multi-view observations.

To this end, we introduce an autoencoding and latent flow framework. Given a 3D Gaussian object $\mathcal{G}$, we first partition it into local Gaussian patches and encode them into a hierarchical latent representation
\begin{equation}
\mathcal{Z}=\left(\mathbf{z}^{g}, \left\{\mathbf{z}^{p}_{m}\right\}_{m=1}^{M}\right),
\end{equation}
where $\mathbf{z}^{g}$ denotes the global latent that captures class-level semantic and holistic shape information, and $\mathbf{z}^{p}_{m}$ denotes the local latent of the $m$-th Gaussian patch. A class-guided rectified flow model is then trained to generate $\mathcal{Z}$ from a simple prior. Finally, the generated latent is decoded into a 3D Gaussian object and further refined through render-feedback supervision.

Formally, the proposed framework consists of a hierarchical Gaussian variational autoencoder (VAE) and a class-guided rectified flow model. The VAE contains an encoder $E$ and a decoder $D$, where the encoder first maps a Gaussian object into a structured latent space,
\begin{equation}
\mathcal{Z}=E(\mathcal{G}),
\end{equation}
the flow model learns to transform a prior latent $\mathcal{Z}_{0}\sim p_{0}$ into a class-conditioned target latent,
\begin{equation}
\hat{\mathcal{Z}}_{1}=\Phi_{\theta}(\mathcal{Z}_{0},c),
\end{equation}
and the decoder reconstructs the final 3D Gaussian object,
\begin{equation}
\hat{\mathcal{G}}=D(\hat{\mathcal{Z}}_{1}).
\end{equation}
For clarity, the proposed framework should be understood through separate training and inference processes. In the first stage, the hierarchical Gaussian VAE learns to encode and decode 3D Gaussian objects as
\begin{equation}
\mathcal{G}\xrightarrow{E}\mathcal{Z}\xrightarrow{D}\hat{\mathcal{G}}.
\end{equation}
After the latent space is established, the second stage trains the class-guided rectified flow model to learn the conditional latent distribution. During inference, generation starts from a prior latent sample:
\begin{equation}
\mathcal{Z}_{0}\xrightarrow{\Phi_{\theta},\,c}\hat{\mathcal{Z}}_{1}\xrightarrow{D}\hat{\mathcal{G}}.
\end{equation}
This formulation more precisely reflects the two-stage nature of the proposed method. It also highlights the key difference from multi-view synthesis or reconstruction-based 3DGS generation: our method directly learns the latent distribution of structured 3D Gaussian objects under class conditions, without requiring an intermediate image generation stage.

\subsection{Hierarchical Gaussian Patch Information Representation}

\begin{figure}[!htbp]
    \centering
    \includegraphics[width=\textwidth]{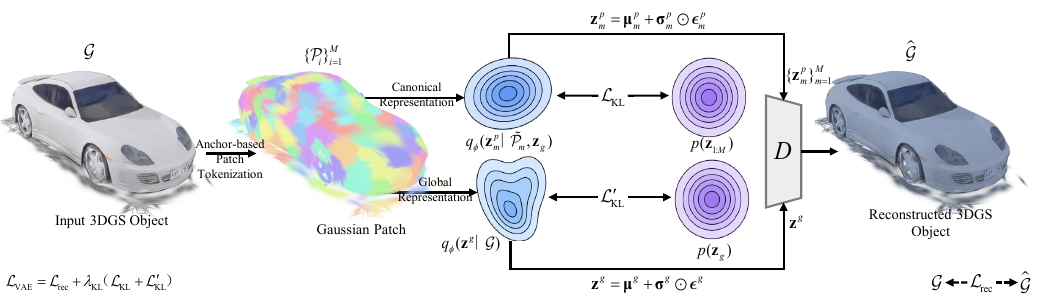}
    \caption{
    Overview of the hierarchical Gaussian patch VAE representation learning stage. This stage establishes a structured and continuous latent space for subsequent class-guided rectified flow generation.
    }
    \label{fig:hierarchical_vae}
\end{figure}

\begin{figure}[!htbp]
  \centering
  \includegraphics[width=0.95\textwidth]{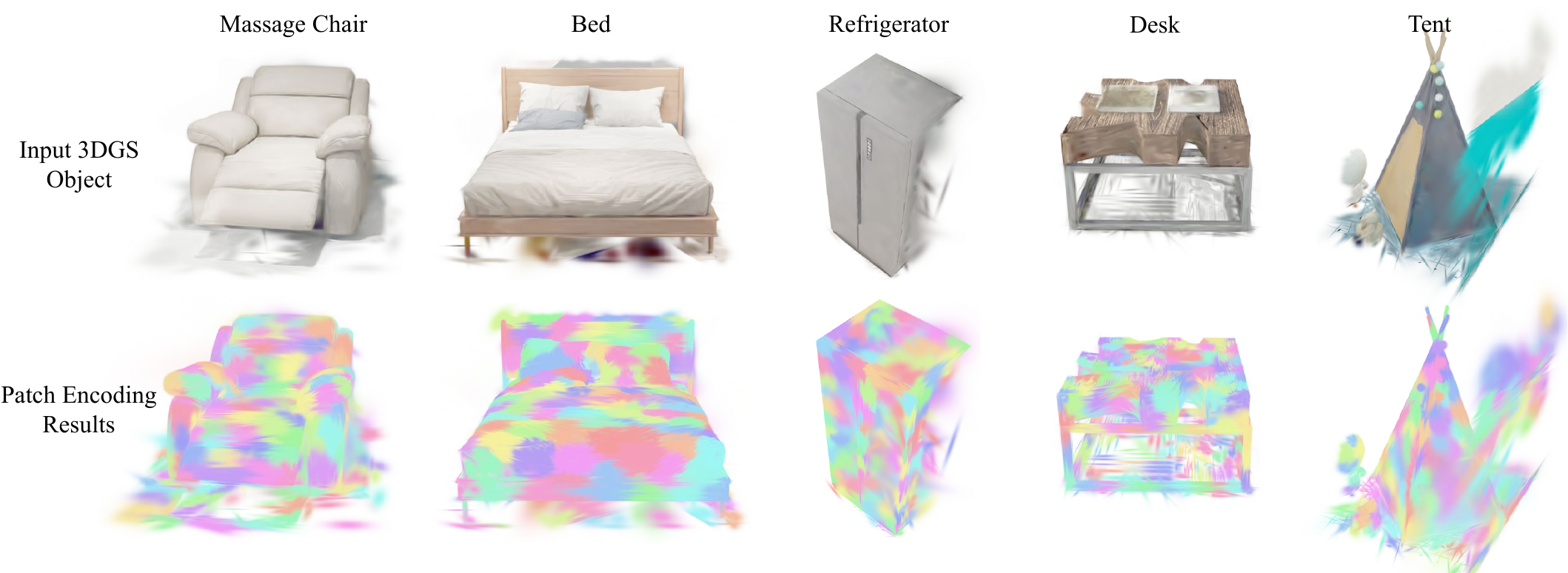}
  \caption{Patch encoding visualization. The first row shows representative input 3DGS objects from different categories, and the second row shows their corresponding patch-encoded representations, where different colors indicate local Gaussian patches.}
  \label{fig:patch_representation}
\end{figure}

A Gaussian object contains an unordered and variable-sized set of primitives whose local density, semantic role, and rendering contribution vary substantially across spatial regions. As illustrated in Fig.~\ref{fig:hierarchical_vae} and visualized in Fig.~\ref{fig:patch_representation}, the irregularity of 3D Gaussian objects makes direct generative modeling difficult. To obtain a more regular representation, we decompose the Gaussian object into a fixed number of local patches and build a hierarchical latent representation over these patches.

Specifically, given $\mathcal{G}=\{\mathbf{g}_i\}_{i=1}^{N}$, we partition it into $M$ local patches, where each patch $\mathcal{P}_{m}$ contains a subset of neighboring Gaussian primitives. The purpose is not merely spatial splitting, but the establishment of stable local units for structured latent modeling.

More concretely, before training, all objects are normalized into a shared canonical object coordinate system, and a fixed set of canonical patch anchors is defined as
\begin{equation}
\bar{\mathcal{C}}=\{\bar{\boldsymbol{\mu}}_{m}\}_{m=1}^{M}.
\end{equation}
These anchors are obtained by applying farthest point sampling (FPS) to the normalized training-set Gaussian centers and are then shared by all objects. Each Gaussian primitive $\mathbf{g}_{i}$ is assigned to its nearest canonical anchor and the $m$-th patch is formed accordingly:
\begin{equation}
\mathcal{G} \rightarrow \left\{\mathcal{P}_{m}\right\}_{m=1}^{M}, \quad
\mathcal{P}_{m}=\{\mathbf{g}_{i}\mid a(i)=m\}, \quad
a(i)=\arg\min_{m}\|\mathbf{x}_{i}-\bar{\boldsymbol{\mu}}_{m}\|_{2}^{2}.
\end{equation}
This anchor-based assignment converts the raw Gaussian set into stable local groups with explicit cross-object correspondence: the same patch index $m$ always refers to the same canonical spatial region. For batched training, the number of primitives inside each patch can be truncated or padded to a preset maximum number $K$, which makes the patch encoder compatible with fixed-size tensor operations. To define a deterministic supervision order inside each local set, Gaussian primitives within each patch are sorted by a canonical local Morton code after coordinate normalization.

For the $m$-th patch, let $\boldsymbol{\mu}_{m}$ denote the patch centroid after assignment and $\mathbf{R}_{m}^{c}$ denote a canonical local orientation. In our implementation, $\boldsymbol{\mu}_{m}$ is computed as the centroid of the Gaussian centers assigned to $\mathcal{P}_{m}$, and it can be regarded as an object-specific offset around the shared anchor $\bar{\boldsymbol{\mu}}_{m}$. The orientation $\mathbf{R}_{m}^{c}$ is obtained by principal component analysis of the patch covariance matrix with deterministic sign disambiguation. Together with a patch extent parameter $\eta_m$ estimated from the spatial spread of the patch, these quantities define the patch frame
\begin{equation}
\mathcal{F}_{m}=\left(\boldsymbol{\mu}_{m},\mathbf{R}^{c}_{m},\eta_m\right).
\end{equation}
The patch frame is used in two ways. During encoding, it defines the canonical coordinate system in which the local Gaussian descriptors are normalized. During decoder training, it also serves as an explicit supervision signal for the global branch, so that the decoder can learn to predict coarse patch frames from the global latent and place generated patches in the global object coordinate system during inference. Each Gaussian primitive inside the patch is transformed into a local canonical coordinate system:
\begin{equation}
\tilde{\mathbf{x}}_{m,j} = (\mathbf{R}_{m}^{c})^\top(\mathbf{x}_{m,j}-\boldsymbol{\mu}_{m}),
\end{equation}
where $\mathbf{x}_{m,j}$ is the original center of the $j$-th Gaussian in patch $\mathcal{P}_m$. Likewise, scale, rotation, opacity, and appearance attributes are normalized or re-parameterized relative to the patch coordinate frame. As a result, each patch is represented by a local Gaussian descriptor set
\begin{equation}
\tilde{\mathcal{P}}_{m}=\left\{\tilde{\mathbf{g}}_{m,j}\right\}_{j=1}^{K_m}, \quad
\tilde{\mathbf{g}}_{m,j}=
\left[
\tilde{\mathbf{x}}_{m,j},
\tilde{\mathbf{s}}_{m,j},
\tilde{\mathbf{r}}_{m,j},
\tilde{o}_{m,j},
\tilde{\mathbf{a}}_{m,j}
\right],
\end{equation}
which reduces the variation caused by global translation, object size, and pose differences. Here $\tilde{\mathbf{s}}_{m,j}$, $\tilde{\mathbf{r}}_{m,j}$, $\tilde{o}_{m,j}$, and $\tilde{\mathbf{a}}_{m,j}$ represent the canonicalized scale, rotation, opacity, and appearance attributes, respectively. This local descriptor preserves geometry, appearance, and rendering-relevant information while removing nuisance variation caused by global translation, scale, and pose differences.

Based on these canonicalized patch descriptors, we construct a hierarchical latent representation. A patch encoder $E_{p}$ maps each local patch into a latent token, then the set of patch tokens is aggregated by a global encoder $E_{g}$ to obtain a global feature, while a local projection module produces patch-level features:
\begin{align}
\mathbf{h}_{m}&=E_{p}(\tilde{\mathcal{P}}_{m}), &
\mathbf{h}^{g}&=E_{g}\left(\left\{\mathbf{h}_{m}\right\}_{m=1}^{M}\right), &
\mathbf{h}^{p}_{m}&=E_{l}(\mathbf{h}_{m},\mathbf{h}^{g}).
\end{align}
To make the latent space generative and continuous, the hierarchical encoder is implemented in a variational form. Instead of directly outputting deterministic latent vectors, the encoder predicts the mean and variance of a posterior distribution
\begin{equation}
q_{\phi}(\mathcal{Z}\mid \mathcal{G})=
q_{\phi}(\mathbf{z}^{g}\mid \mathcal{G})
\prod_{m=1}^{M}
q_{\phi}(\mathbf{z}^{p}_{m}\mid \tilde{\mathcal{P}}_{m},\mathbf{z}^{g}),
\end{equation}
where the global and local posterior factors are modeled as diagonal Gaussian distributions:
\begin{equation}
\begin{aligned}
q_{\phi}(\mathbf{z}^{g}\mid \mathcal{G})
&=
\mathcal{N}(\boldsymbol{\mu}^{g},\mathrm{diag}((\boldsymbol{\sigma}^{g})^{2})), 
    q_{\phi}(\mathbf{z}^{p}_{m}\mid \tilde{\mathcal{P}}_{m},\mathbf{z}^{g})
&=
\mathcal{N}(\boldsymbol{\mu}^{p}_{m},\mathrm{diag}((\boldsymbol{\sigma}^{p}_{m})^{2})).
\end{aligned}
\end{equation}
Variational heads attached to the global feature $\mathbf{h}^{g}$ predict $(\boldsymbol{\mu}^{g},\boldsymbol{\sigma}^{g})$, and variational heads attached to each local feature $\mathbf{h}^{p}_{m}$ predict $(\boldsymbol{\mu}^{p}_{m},\boldsymbol{\sigma}^{p}_{m})$. Latent samples are obtained by the reparameterization trick:
\begin{equation}
\mathbf{z}^{g}=\boldsymbol{\mu}^{g}+\boldsymbol{\sigma}^{g}\odot \boldsymbol{\epsilon}^{g},
\quad
\mathbf{z}^{p}_{m}=\boldsymbol{\mu}^{p}_{m}+\boldsymbol{\sigma}^{p}_{m}\odot \boldsymbol{\epsilon}^{p}_{m},
\end{equation}
where $\boldsymbol{\epsilon}^{g}$ and $\boldsymbol{\epsilon}^{p}_{m}$ are drawn from standard normal distributions. The prior over the hierarchical latent is defined as
\begin{equation}
p_{0}(\mathcal{Z})=
\mathcal{N}(\mathbf{z}^{g};\mathbf{0},\mathbf{I})
\prod_{m=1}^{M}
\mathcal{N}(\mathbf{z}^{p}_{m};\mathbf{0},\mathbf{I}).
\end{equation}
Therefore, KL regularization is imposed on both the global latent and all local patch latents, which encourages the learned hierarchical latent space to remain compatible with prior sampling during the subsequent flow generation stage.

The final hierarchical representation is then written as
\begin{equation}
\mathcal{Z}=\left(\mathbf{z}^{g}, \left\{\mathbf{z}^{p}_{m}\right\}_{m=1}^{M}\right).
\end{equation}
Although the patch frames $\{\mathcal{F}_{m}\}_{m=1}^{M}$ are not treated as stochastic latent variables, they are retained as deterministic structural targets during representation learning. This design separates object-level stochastic generation from spatial layout reconstruction: the global latent $\mathbf{z}^{g}$ captures the distribution of coarse object structure, while the decoder is trained to recover patch frames that provide spatial anchors for local Gaussian patches.

The implementation details of the hierarchical encoder are summarized in the experimental setup.

\begin{figure}[!htbp]
  \centering
  \includegraphics[width=\textwidth]{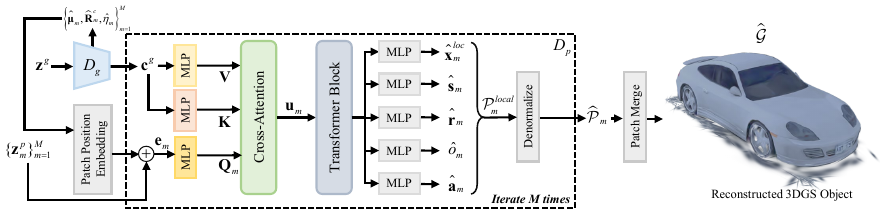}
  \caption{Hierarchical decoder architecture. The global branch derives object-level context and coarse patch frames from the global latent $\mathbf{z}^{g}$, while the local branch reconstructs each canonical Gaussian patch from the local latent $\mathbf{z}^{p}_{m}$ under cross-attention conditioning from the global context. The decoded local Gaussian attributes are denormalized with the predicted patch frame and then merged into the final 3D Gaussian object.}
  \label{fig:hierarchical_decoder}
\end{figure}

As illustrated in Fig.~\ref{fig:hierarchical_decoder}, the decoder mirrors the proposed hierarchical representation. Instead of decoding the whole Gaussian object from a single latent code, it first uses a global decoding branch to infer object-level semantic context and a coarse patch layout from the global latent $\mathbf{z}^{g}$. The global branch is written as
\begin{equation}
\mathbf{c}^{g},\left\{ \hat{\boldsymbol{\mu}}_{m}, \hat{\mathbf{R}}^{c}_{m}, \hat{\eta}_{m}\right\}_{m=1}^{M}
=D_{g}(\mathbf{z}^{g}),
\end{equation}
where $\mathbf{c}^{g}$ denotes the decoded global context, $\hat{\boldsymbol{\mu}}_{m}$ and $\hat{\mathbf{R}}^{c}_{m}$ denote the predicted center and canonical orientation of the $m$-th patch, and $\hat{\eta}_{m}$ denotes an optional patch-scale or extent parameter used for denormalization. The predicted center is initialized around the shared canonical anchor $\bar{\boldsymbol{\mu}}_{m}$, so the global branch only needs to learn object-specific layout residuals. During reconstruction training, the predicted patch frames are supervised by the patch frames obtained from the encoder-side anchor assignment; during generation, these predicted frames provide the spatial support required to place local patches back into the global object coordinate system.

For each local latent $\mathbf{z}^{p}_{m}$, a patch-position embedding based on the shared anchor is added to preserve the spatial role of the patch:
\begin{equation}
\mathbf{e}_{m}=\mathbf{z}^{p}_{m}+\mathrm{PE}(\bar{\boldsymbol{\mu}}_{m}).
\end{equation}
The local decoder then lets the position-aware local token query the global context through cross-attention. Specifically, the local token produces the query, while the global context produces keys and values:
\begin{equation}
\mathbf{Q}_{m}=W_{Q}\mathbf{e}_{m}, \quad
\mathbf{K}=W_{K}\mathbf{C}^{g}, \quad
\mathbf{V}=W_{V}\mathbf{C}^{g},
\end{equation}
\begin{equation}
\mathbf{u}_{m}
=
\mathrm{Softmax}
\left(
\frac{\mathbf{Q}_{m}\mathbf{K}^{\top}}{\sqrt{d_{k}}}
\right)\mathbf{V},
\end{equation}
where $\mathbf{C}^{g}$ denotes the global context tokens derived from $\mathbf{c}^{g}$ and $d_k$ is the key dimension. The attended local feature is then passed to several attribute-specific prediction heads:
\begin{equation}
\left[
\hat{\mathbf{x}}^{\mathrm{loc}}_{m},
\hat{\mathbf{s}}_{m},
\hat{\mathbf{r}}_{m},
\hat{\mathbf{o}}_{m},
\hat{\mathbf{a}}_{m}
\right]
=
\left[
H_{x}(\mathbf{u}_{m}),
H_{s}(\mathbf{u}_{m}),
H_{r}(\mathbf{u}_{m}),
H_{o}(\mathbf{u}_{m}),
H_{a}(\mathbf{u}_{m})
\right],
\end{equation}
where the five heads predict the local Gaussian center offsets, scales, rotations, opacities, and appearance attributes of the $m$-th patch, respectively. This multi-head design allows the decoder to model different Gaussian attributes with shared structural context but attribute-specific output projections.

The predicted attributes form a canonical local Gaussian patch
\begin{equation}
\hat{\mathcal{P}}^{\mathrm{loc}}_{m}
=
\left\{
\left(
\hat{\mathbf{x}}^{\mathrm{loc}}_{m,j},
\hat{\mathbf{s}}_{m,j},
\hat{\mathbf{r}}_{m,j},
\hat{o}_{m,j},
\hat{\mathbf{a}}_{m,j}
\right)
\right\}_{j=1}^{K}.
\end{equation}
It is then denormalized and mapped back to the global coordinate system using the predicted patch frame:
\begin{equation}
\hat{\mathcal{P}}_{m}
=
T^{-1}_{m}
\left(
\hat{\mathcal{P}}^{\mathrm{loc}}_{m};
\hat{\boldsymbol{\mu}}_{m},
\hat{\mathbf{R}}^{c}_{m},
\hat{\eta}_{m}
\right).
\end{equation}
Finally, all decoded patches are merged into the reconstructed Gaussian object. This hierarchical decoder keeps local Gaussian details conditioned on the global object identity, while the predicted patch frames and position embeddings provide spatial correspondence for assembling a coherent 3D Gaussian object.

The proposed representation has three advantages. First, it converts an unordered Gaussian set into a structured token sequence with explicit local correspondence. Second, it disentangles global class semantics from local geometric-appearance details. Third, it creates a regular latent domain that is more suitable for direct flow-based generative modeling than raw Gaussian primitives or a single global latent code.

\subsection{Class-Guided Structure-Aware Rectified Flow}
After obtaining the hierarchical latent representation, we perform direct class-guided generation in latent space by rectified flow. For each training object, let
\begin{equation}
\mathcal{Z}_{1}=\left(\mathbf{z}^{g}_{1},\left\{\mathbf{z}^{p}_{1,m}\right\}_{m=1}^{M}\right), \quad
\mathcal{Z}_{0}=\left(\mathbf{z}^{g}_{0},\left\{\mathbf{z}^{p}_{0,m}\right\}_{m=1}^{M}\right)
\end{equation}
denote the target latent encoded from a real 3D Gaussian object and a prior sample drawn from a standard Gaussian distribution, respectively. For time $t \in [0,1]$, we define the interpolated latent state as
\begin{equation}
\mathbf{z}^{g}_{t}=(1-t)\mathbf{z}^{g}_{0}+t\mathbf{z}^{g}_{1}, \quad
\mathbf{z}^{p}_{t,m}=(1-t)\mathbf{z}^{p}_{0,m}+t\mathbf{z}^{p}_{1,m}.
\end{equation}

For convenience, the interpolated hierarchical latent can be written as
\begin{equation}
\mathcal{Z}_{t}=\left(\mathbf{z}^{g}_{t},\left\{\mathbf{z}^{p}_{t,m}\right\}_{m=1}^{M}\right),
\end{equation}
and the target velocities of the two branches are correspondingly defined as
\begin{equation}
\mathbf{v}^{g\ast}=\mathbf{z}^{g}_{1}-\mathbf{z}^{g}_{0}, \quad
\mathbf{v}^{p\ast}_{m}=\mathbf{z}^{p}_{1,m}-\mathbf{z}^{p}_{0,m}.
\end{equation}

To incorporate class information, the class label $c$ is embedded into a condition vector $\mathbf{e}_{c}$. To preserve structural correspondence among local patches, we further attach a patch-position encoding to each local latent:
\begin{equation}
\hat{\mathbf{z}}^{p}_{t,m}=\mathbf{z}^{p}_{t,m}+\mathrm{PE}(\bar{\boldsymbol{\mu}}_{m}),
\end{equation}
where $\bar{\boldsymbol{\mu}}_{m}$ is the $m$-th shared canonical patch anchor and $\mathrm{PE}(\cdot)$ denotes the positional embedding. Because the anchors are fixed and shared by all objects, the same position encoding is available during both training and inference. This design enables the flow model to distinguish patch roles according to canonical object layout rather than treating all local tokens as exchangeable latent vectors.

The proposed structure-aware rectified flow jointly predicts global and local velocities through a global-local coupled velocity network. The time variable $t$ is first mapped to a time embedding:
\begin{equation}
\mathbf{e}_{t}=\mathrm{MLP}_{t}(\mathrm{SinEmb}(t)).
\end{equation}
For the local flow branch, each patch token is initialized by combining the position-aware local latent, the time embedding, and the class embedding:
\begin{equation}
\mathbf{h}^{p,0}_{t,m}
=
\mathrm{FFN}^{p}_{\mathrm{in}}
\left(
\left[
\hat{\mathbf{z}}^{p}_{t,m},
\mathbf{e}_{t},
\mathbf{e}_{c}
\right]
\right).
\end{equation}
The set of local tokens is then processed by stacked self-attention blocks to model structural dependencies among Gaussian patches:
\begin{equation}
\left\{
\mathbf{h}^{p}_{t,m}
\right\}_{m=1}^{M}
=
\mathrm{SelfAttn}^{p}_{\theta}
\left(
\left\{
\mathbf{h}^{p,0}_{t,m}
\right\}_{m=1}^{M}
\right).
\end{equation}
The resulting local tokens are further aggregated into a local structural context:
\begin{equation}
\mathbf{s}^{p}_{t}
=
\mathrm{Agg}
\left(
\left\{
\mathbf{h}^{p}_{t,m}
\right\}_{m=1}^{M}
\right),
\end{equation}
where $\mathrm{Agg}(\cdot)$ is implemented as attention pooling.

\begin{figure}[!htbp]
  \centering
  \includegraphics[width=0.4\textwidth]{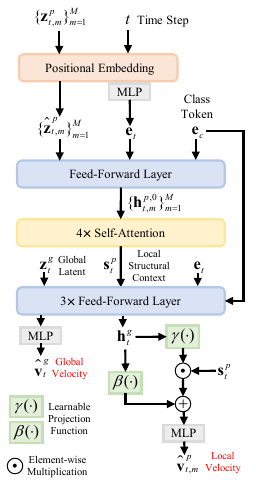}
  \caption{Illustration of the class-guided structure-aware rectified flow model. The local flow branch extracts patch-level structural tokens, while the global flow branch fuses the global latent, local structural context, time embedding, and class embedding to predict the global velocity and modulate local velocity prediction.}
  \label{fig:structure_aware_flow}
\end{figure}

As shown in Fig.~\ref{fig:structure_aware_flow}, the global flow branch fuses the current global latent, the aggregated local structural context, the time embedding, and the class embedding:
\begin{equation}
\mathbf{h}^{g}_{t}
=
\mathrm{FFN}^{g}_{\theta}
\left(
\left[
\mathbf{z}^{g}_{t},
\mathbf{s}^{p}_{t},
\mathbf{e}_{t},
\mathbf{e}_{c}
\right]
\right).
\end{equation}
The global velocity is then predicted by an output multilayer perceptron:
\begin{equation}
\hat{\mathbf{v}}^{g}_{t}
=
\mathrm{MLP}^{g}_{\mathrm{out}}
\left(
\mathbf{h}^{g}_{t}
\right).
\end{equation}
To inject object-level guidance back into local velocity prediction, the global hidden feature $\mathbf{h}^{g}_{t}$ is used to generate two adaptive modulation terms, $\gamma(\mathbf{h}^{g}_{t})$ and $\beta(\mathbf{h}^{g}_{t})$. Each patch-level feature is modulated as
\begin{equation}
\tilde{\mathbf{h}}^{p}_{t,m}
=
\gamma(\mathbf{h}^{g}_{t})
\odot
\mathbf{h}^{p}_{t,m}
+
\beta(\mathbf{h}^{g}_{t}),
\end{equation}
where $\odot$ denotes element-wise multiplication. The local flow branch finally predicts the patch-level velocity:
\begin{equation}
\hat{\mathbf{v}}^{p}_{t,m}
=
\mathrm{MLP}^{p}_{\mathrm{out}}
\left(
\tilde{\mathbf{h}}^{p}_{t,m}
\right),
\quad m=1,\dots,M.
\end{equation}
Therefore, the complete velocity network can be written as
\begin{equation}
F_{\theta}
\left(
\mathcal{Z}_{t},
t,
\mathbf{e}_{c}
\right)
=
\left(
\hat{\mathbf{v}}^{g}_{t},
\left\{
\hat{\mathbf{v}}^{p}_{t,m}
\right\}_{m=1}^{M}
\right).
\end{equation}
Different from vanilla rectified flow with a single latent variable, our model explicitly couples global and local velocity prediction. The local branch captures patch-level geometry, appearance, and structural variation, while the global branch captures class-level semantics and object-level layout. The attention-pooled local structural context allows the global branch to absorb patch-level evidence, and the adaptive modulation enables the global object state to guide local patch generation.

Moreover, not all Gaussian patches contribute equally to the final object quality. Patches with higher opacity, richer geometry, or stronger visible contribution are more important than weakly contributing patches. We therefore introduce density-aware velocity weighting:
\begin{equation}
\bar{o}_{m}=\frac{1}{K_m}\sum_{j=1}^{K_m} o_{m,j}, \quad
\rho_{m}=\frac{K_m}{V_m+\epsilon},
\end{equation}
where $\bar{o}_{m}$ is the mean opacity, $\rho_m$ is the local primitive density, and $V_m$ denotes the occupied spatial volume of patch $\mathcal{P}_{m}$. Based on these statistics, the patch importance weight is defined as
\begin{equation}
w_m=
\frac{\exp(\alpha \bar{o}_{m}+\beta \rho_{m}+\gamma \nu_{m})}
{\sum_{m'=1}^{M}\exp(\alpha \bar{o}_{m'}+\beta \rho_{m'}+\gamma \nu_{m'})},
\end{equation}
where $\nu_{m}$ is the average alpha-composited visibility score of patch $\mathcal{P}_m$ estimated over the rendered supervision views. This formulation encourages the model to focus more on patches that have larger visual or structural impact on the final object.

With the above definitions, the global and local velocity losses are
\begin{align}
\mathcal{L}_{g} &= \left\|\hat{\mathbf{v}}^{g}_{t}-\mathbf{v}^{g\ast}\right\|_{2}^{2}, \\
\mathcal{L}_{p} &= \sum_{m=1}^{M} w_{m} \left\|\hat{\mathbf{v}}^{p}_{t,m}-\mathbf{v}^{p\ast}_{m}\right\|_{2}^{2},
\end{align}
and the final structure-aware flow loss is
\begin{equation}
\mathcal{L}_{\mathrm{flow}}=
\lambda_{g}\mathcal{L}_{g}
\;+\;
\lambda_{\ell}\mathcal{L}_{p},
\end{equation}
where $\lambda_{g}$ and $\lambda_{\ell}$ balance the global and local velocity objectives. Compared with uniform latent supervision, this design better reflects the heterogeneous contribution of different Gaussian regions to object structure and appearance.

\begin{figure}[!t]
  \centering
  \includegraphics[width=\textwidth]{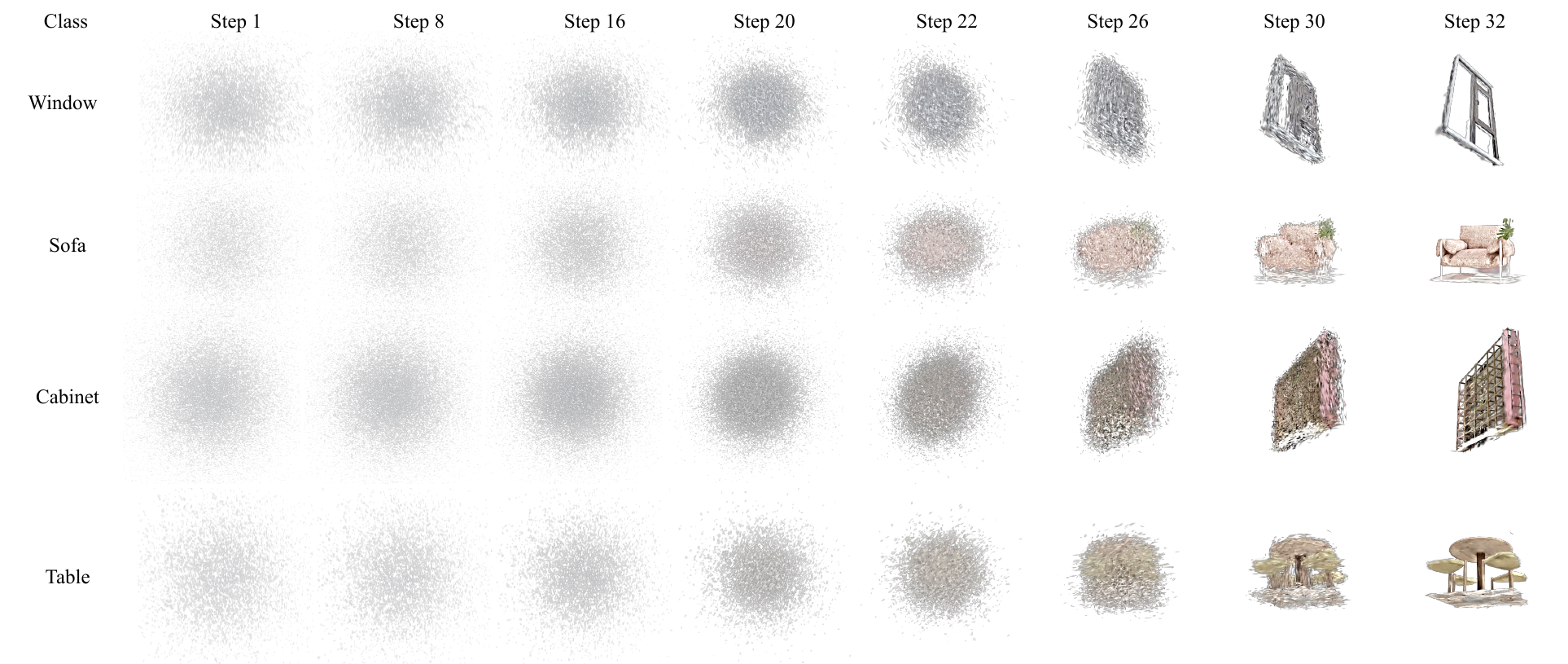}
  \caption{Visualization of the class-guided rectified-flow generation trajectory. Each row corresponds to a class condition, and each column shows the decoded 3D Gaussian object at a different integration step. The generated object progressively evolves from noisy Gaussian primitives to a coherent class-specific structure.}
  \label{fig:generation_trajectory}
\end{figure}

Fig.~\ref{fig:generation_trajectory} visualizes the class-guided generation trajectory of the proposed rectified flow. Starting from the Gaussian prior latent $\mathcal{Z}_{0}$, the decoded 3D Gaussian object gradually evolves from noisy primitives into a coherent class-specific structure as the Euler integration proceeds.

Overall, the proposed flow model directly learns the class-conditioned distribution of structured 3D Gaussian latents, which fundamentally differs from methods that first synthesize multi-view images and then reconstruct 3DGS objects.

\subsection{Render-Feedback Flow Learning}
Although the rectified flow is trained in latent space, the ultimate quality of a generated 3D Gaussian object is determined by its decoded geometry and multi-view appearance. A latent that matches the target distribution in Euclidean space does not necessarily correspond to a visually coherent 3DGS object after decoding. We therefore introduce render-feedback flow learning to connect latent generation with the rendering behavior of decoded Gaussian objects.

Let $\hat{\mathcal{Z}}_{1}$ denote the predicted terminal latent obtained by integrating the learned flow from a prior sample. A Gaussian decoder $D$ reconstructs the corresponding 3D object:
\begin{equation}
\hat{\mathcal{G}}=D(\hat{\mathcal{Z}}_{1}), \quad
\mathcal{G}^{\ast}=D(\mathcal{Z}_{1}),
\end{equation}
where $\mathcal{G}^{\ast}$ is the reconstructed object from the ground-truth latent $\mathcal{Z}_{1}$ (or equivalently the original training Gaussian object when direct decoding consistency is enforced). We sample a fixed set of camera views $\{\pi_k\}_{k=1}^{K_v}$ and render both objects:
\begin{equation}
\hat{\mathbf{I}}_{k}=\mathcal{R}(\hat{\mathcal{G}},\pi_k), \quad
\mathbf{I}^{\ast}_{k}=\mathcal{R}(\mathcal{G}^{\ast},\pi_k).
\end{equation}
The camera set $\{\pi_k\}_{k=1}^{K_v}$ is sampled uniformly on the upper viewing hemisphere. These rendered views serve as image-space probes for evaluating whether the generated latent remains consistent after decoding and rendering.

The render-feedback loss is then defined as
\begin{equation}
\mathcal{L}_{\mathrm{render}}=
\frac{1}{K_v}\sum_{k=1}^{K_v}
\left[
\lambda_{1}\left\|\hat{\mathbf{I}}_{k}-\mathbf{I}^{\ast}_{k}\right\|_{1}
\;+\;
\lambda_{s}\mathcal{L}_{\mathrm{ssim}}(\hat{\mathbf{I}}_{k},\mathbf{I}^{\ast}_{k})
\right],
\end{equation}
where $\mathcal{L}_{\mathrm{ssim}}$ denotes a structural image similarity loss. The main role of render-feedback is to provide decoder-aware and renderer-aware correction so that the learned latent flow is not only distributionally plausible but also visually valid.

This strategy is particularly important for 3DGS generation because local Gaussian errors may be weak in latent space but amplified after rendering. By imposing multi-view render supervision, the model is encouraged to produce objects with coherent geometry, sharper local appearance, and improved view consistency. To reduce training cost, render-feedback is applied periodically during flow optimization.

In other words, the proposed training scheme does not stop at latent alignment. It explicitly regularizes the compatibility between the generated latent, the hierarchical Gaussian decoder, and the downstream renderer, which is especially important for 3DGS because its perceptual quality is highly sensitive to splatting behavior, opacity accumulation, and multi-view consistency.

\subsection{Training Objective and Inference Procedure}
The full training objective consists of three parts: a hierarchical representation learning objective, a structure-aware rectified flow objective, and a render-feedback objective. First, the encoder-decoder pair is trained to reconstruct the Gaussian object from the hierarchical latent representation. The reconstruction loss can be formulated as
\begin{equation}
\mathcal{L}_{\mathrm{rec}}=
\lambda_{x}\mathcal{L}_{x}+
\lambda_{\Sigma}\mathcal{L}_{\Sigma}+
\lambda_{o}\mathcal{L}_{o}+
\lambda_{a}\mathcal{L}_{a}+
\lambda_{f}\mathcal{L}_{\mathrm{frame}},
\end{equation}
where $\mathcal{L}_{x}$, $\mathcal{L}_{\Sigma}$, $\mathcal{L}_{o}$, and $\mathcal{L}_{a}$ supervise Gaussian center, covariance, opacity, and appearance reconstruction, respectively. The additional term $\mathcal{L}_{\mathrm{frame}}$ supervises the coarse patch frames predicted by the global decoder:
\begin{equation}
\mathcal{L}_{\mathrm{frame}}
=
\frac{1}{M}\sum_{m=1}^{M}
\left(
\left\|\hat{\boldsymbol{\mu}}_{m}-\boldsymbol{\mu}_{m}\right\|_{1}
+
\lambda_{R}\left\|\hat{\mathbf{R}}^{c}_{m}-\mathbf{R}^{c}_{m}\right\|_{F}^{2}
+
\lambda_{\eta}\left\|\hat{\eta}_{m}-\eta_{m}\right\|_{1}
\right).
\end{equation}
Here $(\boldsymbol{\mu}_{m},\mathbf{R}^{c}_{m},\eta_m)$ are the encoder-side patch frames obtained during patch partition and canonicalization. This supervision makes the global latent responsible not only for object-level semantics but also for the coarse spatial anchors required by the local decoder. The reconstruction loss types and their weights are summarized in the experimental setup.

In our framework, the hierarchical representation is trained as a variational autoencoder in the first stage. Therefore, besides reconstruction supervision, a latent regularization term is imposed:
\begin{equation}
\mathcal{L}_{\mathrm{VAE}}=\mathcal{L}_{\mathrm{rec}}+\lambda_{\mathrm{KL}}\mathcal{L}_{\mathrm{KL}}.
\end{equation}
Here $\mathcal{L}_{\mathrm{KL}}$ regularizes the latent posterior toward the standard normal prior defined above. It is computed over both the global latent and the local patch latents:
\begin{equation}
\mathcal{L}_{\mathrm{KL}}=
D_{\mathrm{KL}}\left(q_{\phi}(\mathbf{z}^{g}\mid \mathcal{G})\| \mathcal{N}(\mathbf{0},\mathbf{I})\right)
+\frac{1}{M}\sum_{m=1}^{M}
D_{\mathrm{KL}}\left(q_{\phi}(\mathbf{z}^{p}_{m}\mid \tilde{\mathcal{P}}_{m},\mathbf{z}^{g})\| \mathcal{N}(\mathbf{0},\mathbf{I})\right).
\end{equation}
This VAE stage is crucial because it establishes a structured and continuous latent space in which the subsequent flow model can operate reliably.

During the second stage, the flow model is trained on top of the latent space learned by the VAE. The total objective of the second stage is
\begin{equation}
\mathcal{L}_{\mathrm{total}}=
\lambda_{\mathrm{flow}}\mathcal{L}_{\mathrm{flow}}
\;+\;
\lambda_{r}\mathcal{L}_{\mathrm{render}}.
\end{equation}
In practice, our training follows a clear two-stage strategy rather than fully joint optimization. In the first stage, the hierarchical Gaussian VAE is trained using $\mathcal{L}_{\mathrm{VAE}}$. In the second stage, the class-guided rectified flow is trained on top of the learned latent space with cosine learning-rate decay. During the second stage, the decoder is fine-tuned with a reduced learning rate, but the second-stage objective does not include the VAE reconstruction or KL terms.

More specifically, in the first stage, the encoder-decoder pair is optimized with $\mathcal{L}_{\mathrm{VAE}}$ until the latent representation becomes stable and reconstructive. In the second stage, the encoded latent samples serve as training targets for the flow model, and $\mathcal{L}_{\mathrm{render}}$ is used as an auxiliary visual correction term to improve decoder compatibility and multi-view consistency of generated objects. The detailed training schedule and loss weights are summarized in the experimental setup.

During inference, only a class label $c$ and a random prior sample are required. We first sample
\begin{equation}
\mathcal{Z}_{0} \sim p_{0}(\mathcal{Z}), \quad
\frac{\mathrm{d}\mathcal{Z}_{t}}{\mathrm{d}t}=F_{\theta}(\mathcal{Z}_{t},t,\mathbf{e}_{c}),
\end{equation}
and then integrate the learned class-guided velocity field from $t=0$ to $t=1$. In practice, this ODE is solved using Euler discretization. After $S$ integration steps, the final latent is obtained as
\begin{equation}
\hat{\mathcal{Z}}_{1}=\mathrm{SolveODE}(F_{\theta},\mathcal{Z}_{0},\mathbf{e}_{c}),
\end{equation}
which is then decoded into a 3D Gaussian object:
\begin{equation}
\hat{\mathcal{G}}=D(\hat{\mathcal{Z}}_{1}),
\end{equation}
which can be directly rendered from arbitrary viewpoints. Since the generation is performed in a structured latent space rather than via per-instance multi-view optimization, the proposed pipeline is more suitable for direct class-guided 3DGS object synthesis.

Algorithm~\ref{alg:training} summarizes the training process of the proposed method.

\begin{algorithm}[t]
\caption{Two-stage training procedure of the proposed class-guided 3DGS generation framework}
\label{alg:training}
\begin{algorithmic}[1]
\Require $\mathcal{D}=\{(\mathcal{G}^{(n)}, c^{(n)})\}_{n=1}^{N_d}$, anchors $\bar{\mathcal{C}}=\{\bar{\boldsymbol{\mu}}_m\}_{m=1}^{M}$, encoder $E_{\phi}$, decoder $D_{\psi}$, flow $F_{\theta}$
\Ensure $\phi,\psi,\theta$
\Statex \textbf{Stage 1: Train hierarchical Gaussian VAE}
\For{$k=1,\dots,K_{\mathrm{VAE}}$}
    \State $\mathcal{B}\sim\mathcal{D}$
    \For{$(\mathcal{G},c)\in\mathcal{B}$}
        \State $\{\mathcal{P}_m\}_{m=1}^{M}\leftarrow \mathrm{AnchorAssign}(\mathcal{G},\bar{\mathcal{C}})$
        \State $\{\mathcal{P}_m\}_{m=1}^{M}\leftarrow \mathrm{MortonSort}(\{\mathcal{P}_m\}_{m=1}^{M})$
        \State $\{\tilde{\mathcal{P}}_m,\mathcal{F}_m\}_{m=1}^{M}\leftarrow \mathrm{Canonicalize}(\{\mathcal{P}_m\}_{m=1}^{M})$
        \State $\mathcal{Z}=E_{\phi}(\{\tilde{\mathcal{P}}_m\}_{m=1}^{M})$
        \State $\hat{\mathcal{G}}=D_{\psi}(\mathcal{Z};\bar{\mathcal{C}})$
    \EndFor
    \State $\mathcal{L}_{\mathrm{VAE}}=\mathcal{L}_{\mathrm{rec}}+\lambda_{\mathrm{KL}}\mathcal{L}_{\mathrm{KL}}$
    \State $(\phi,\psi)\leftarrow(\phi,\psi)-\alpha_{\mathrm{VAE}}\nabla_{\phi,\psi}\mathcal{L}_{\mathrm{VAE}}$
\EndFor
\Statex \textbf{Stage 2: Train class-guided rectified flow}
\For{$k=1,\dots,K_{\mathrm{flow}}$}
    \State $\mathcal{B}\sim\mathcal{D},\quad t\sim\mathcal{U}(0,1)$
    \State $\mathcal{Z}_{1}=E_{\phi}(\mathcal{G}),\quad \mathcal{Z}_{0}\sim p_0(\mathcal{Z}),\quad \mathcal{Z}_{t}=(1-t)\mathcal{Z}_{0}+t\mathcal{Z}_{1}$
    \State $\hat{\mathcal{V}}_{t}=F_{\theta}(\mathcal{Z}_{t},t,\mathbf{e}_{c},\{\mathrm{PE}(\bar{\boldsymbol{\mu}}_m)\}_{m=1}^{M})$
    \State $\mathcal{L}_{\mathrm{flow}}\leftarrow \mathrm{VelocityLoss}(\hat{\mathcal{V}}_{t},\mathcal{Z}_{1}-\mathcal{Z}_{0},\{w_m\}_{m=1}^{M})$
    \State $\hat{\mathcal{Z}}_{1}=\mathrm{Euler}_{S}(F_{\theta},\mathcal{Z}_{0},\mathbf{e}_{c}),\quad \hat{\mathcal{G}}=D_{\psi}(\hat{\mathcal{Z}}_{1};\bar{\mathcal{C}})$
    \State $\mathcal{L}_{\mathrm{render}}\leftarrow \mathrm{RenderLoss}(\hat{\mathcal{G}},\mathcal{G}^{\ast},\{\pi_j\}_{j=1}^{K_v})$
    \State $\mathcal{L}_{\mathrm{total}}=\lambda_{\mathrm{flow}}\mathcal{L}_{\mathrm{flow}}+\lambda_r\mathcal{L}_{\mathrm{render}}$
    \State $(\theta,\psi)\leftarrow(\theta,\psi)-\alpha_{\mathrm{flow}}\nabla_{\theta,\psi}\mathcal{L}_{\mathrm{total}}$
\EndFor
\end{algorithmic}
\end{algorithm}

\section{Experiments}
\subsection{Experimental Setup}
To evaluate the proposed class-guided 3D Gaussian generation framework, we construct a mixed public dataset from several widely used 3D repositories and indoor 3DGS resources, including ShapeNetCore~\citep{chang2015shapenet}, Objaverse~\citep{deitke2023objaverse}, Amazon Berkeley Objects (ABO)~\citep{collins2022abo}, 3D-FUTURE~\citep{fu20213dfuture}, and InteriorGS~\citep{interiorgs2025}. These datasets provide complementary object and scene distributions. ShapeNetCore contains clean and category-organized CAD models, Objaverse provides large-scale and diverse 3D assets, ABO contributes real product-level objects with high-quality textures and physically based materials, and 3D-FUTURE offers high-quality furniture assets with detailed geometry and appearance. InteriorGS further complements these object-centric sources with semantically labeled indoor 3DGS scenes and object instances, which improves the coverage of indoor categories and scene-derived Gaussian structures.

We select common object categories that appear consistently across these datasets and merge semantically similar labels into a unified category taxonomy. The final dataset contains 34 object categories, covering furniture, daily objects, electronic devices, and transportation-related objects. Categories with too few valid samples, incomplete geometry, missing object-level structure, or severe texture artifacts are removed. This filtering ensures that each retained category has sufficient intra-class diversity while preserving stable class semantics for conditional generation.

All collected 3D assets are converted into a unified object-level 3DGS representation before training. For each object, we first normalize the mesh or scene asset into a canonical coordinate system by centering it at the origin and scaling it into a unit bounding volume. Objects with disconnected fragments, extremely sparse geometry, or invalid material files are discarded. The normalized object is then rendered or optimized into a 3D Gaussian object with learnable position, covariance, opacity, and appearance attributes. After 3DGS conversion, each object is further processed by the proposed patch construction pipeline, where Gaussian primitives are assigned to shared canonical patch anchors and represented in local canonical coordinates.

The dataset is split at the object level into training, validation, and testing subsets. The split is performed independently within each category to avoid category imbalance and to ensure that no object instance appears in more than one subset. Unless otherwise specified, all models are trained only with category labels as conditions, without using text prompts or image inputs during generation.

For comparison, we evaluate the proposed method against two groups of representative baselines. The first group consists of published 3DGS generative models, including GaussianCube~\citep{zhang2024gaussiancube}, DiffGS~\citep{zhou2024diffgs}, and SplatFlow~\citep{go2025splatflow}. GaussianCube represents a grid-structured Gaussian generation baseline, DiffGS represents a diffusion-based functional Gaussian generation baseline, and SplatFlow represents a rectified-flow-based 3DGS synthesis baseline. These methods are selected to compare different strategies for modeling the unordered and structure-sensitive nature of 3D Gaussian primitives.

The second group consists of direct 3DGS adaptations of representative image-space generative backbones, including DiT-3DGS, U-ViT-3DGS, and Vanilla RF-3DGS. These baselines lift 2D image diffusion or flow architectures to the 3DGS latent dimension by flattening Gaussian patch latents into token sequences or dense latent grids, while keeping the original backbone design largely unchanged. This comparison highlights the difference between naively extending image generative models to 3DGS representations and explicitly modeling the structured global-local distribution of 3D Gaussian objects.

We evaluate all methods from four aspects: geometry quality, rendering quality, distribution similarity, and multi-view consistency. Geometry quality measures whether the generated object has plausible 3D structure and complete shape. Rendering quality evaluates the visual fidelity of rendered views. Distribution similarity measures how well generated samples match the real object distribution within each category. Multi-view consistency evaluates whether the same generated object remains coherent under different camera views. Efficiency is also reported in terms of inference time and memory usage.

Table~\ref{tab:exp_setup} summarizes the implementation details and hyperparameter settings used in our experiments.

\begin{table}[t]
\caption{Implementation details and hyperparameter settings of the proposed method.}
\label{tab:exp_setup}
\scriptsize
\begin{tabular*}{\textwidth}{@{\extracolsep{\fill}}p{0.26\textwidth}p{0.66\textwidth}@{}}
\toprule
Component & Setting \\
\midrule
Patch anchors & FPS on normalized Gaussian centers \\
Patch ordering & Morton order after local coordinate normalization \\
Latent dimensions & $d_g=256$, $d_p=128$ \\
Condition embeddings & $\dim(\mathbf{e}_c)=128$; $\mathrm{PE}(\bar{\boldsymbol{\mu}}_m)$ projected to 128 dimensions \\
Density-aware weights & $\alpha=1.0$, $\beta=0.5$, $\gamma=0.5$ \\
Flow loss weights & $\lambda_g=1.0$, $\lambda_{\ell}=2.0$ \\
Render-feedback views & $K_v=4$; azimuth $[0,2\pi)$; elevation $[\pi/12,\pi/3]$ \\
Render-feedback loss & $\lambda_1=1.0$, $\lambda_s=0.2$; applied every two flow iterations \\
Reconstruction loss weights & $\lambda_x=1.0$, $\lambda_{\Sigma}=1.0$, $\lambda_o=0.5$, $\lambda_a=1.0$, $\lambda_f=0.5$, $\lambda_R=0.1$, $\lambda_{\eta}=0.1$ \\
KL regularization & $\lambda_{\mathrm{KL}}=10^{-4}$ \\
Stage 1 training & 200 epochs; AdamW; learning rate $1\times10^{-4}$; batch size 8 \\
Stage 2 training & 300 epochs; AdamW; learning rate $2\times10^{-4}$; batch size 16 \\
Stage 2 loss weights & $\lambda_{\mathrm{flow}}=1.0$, $\lambda_r=0.5$ \\
Inference & Euler solver; $S=32$ integration steps \\
\bottomrule
\end{tabular*}
\end{table}

\subsection{Evaluation of Hierarchical Gaussian VAE}
Before comparing the full generative pipeline with state-of-the-art methods, we first evaluate whether the proposed hierarchical Gaussian VAE provides a faithful and structured latent space for 3DGS objects. This evaluation is important because the subsequent rectified flow model is trained entirely in the learned latent space. If the autoencoder loses local geometry, patch layout, or rendering-sensitive attributes, the generated latent samples cannot be reliably decoded into high-quality 3D Gaussian objects.

Table~\ref{tab:vae_comparison} compares the proposed hierarchical VAE with several representation-learning variants. The global VAE compresses the whole 3DGS object into a single latent code, while the flat patch VAE encodes local patches without global-local hierarchical coupling. The variant without canonicalization directly encodes raw patch coordinates, and the variant without frame supervision removes the explicit patch-frame reconstruction objective. The proposed hierarchical VAE achieves the best reconstruction and rendering performance, showing that canonical patch encoding, global-local latent decomposition, and patch-frame supervision are all important for preserving 3DGS structure.

\begin{table}[b]
\caption{Comparison of different 3DGS autoencoding representations. Best results are shown in bold.}
\label{tab:vae_comparison}
\centering
\scriptsize
\setlength{\tabcolsep}{5pt}
\begin{tabular*}{\textwidth}{@{\extracolsep{\fill}}lcccccc@{}}
\toprule
\multirow{2}{*}{Method}
& \multicolumn{2}{c}{Geometry}
& \multicolumn{2}{c}{Rendering}
& \multicolumn{2}{c}{Structure} \\
\cmidrule(lr){2-3}
\cmidrule(lr){4-5}
\cmidrule(lr){6-7}
& CD $\downarrow$ & F@1 $\uparrow$
& LPIPS $\downarrow$ & SSIM $\uparrow$
& FE $\downarrow$ & PE $\downarrow$ \\
\midrule
Global VAE & 3.94 & 57.8 & 0.205 & 0.733 & 0.081 & 0.126 \\
Flat patch VAE & 3.31 & 62.4 & 0.183 & 0.756 & 0.064 & 0.103 \\
w/o canonicalization & 3.12 & 64.8 & 0.174 & 0.768 & 0.058 & 0.097 \\
w/o frame supervision & 2.95 & 66.1 & 0.166 & 0.779 & 0.071 & 0.089 \\
Ours & \textbf{2.41} & \textbf{72.5} & \textbf{0.128} & \textbf{0.826} & \textbf{0.039} & \textbf{0.061} \\
\bottomrule
\end{tabular*}
\par\vspace{0.3em}
\parbox{0.92\textwidth}{\footnotesize \textit{Note}: CD: Chamfer Distance ($10^{-3}$); F@1: F-score at 1\% distance threshold (\%); LPIPS: Learned Perceptual Image Patch Similarity; SSIM: Structural Similarity; FE: patch-frame error; PE: patch-position error.}
\end{table}

\begin{figure}[thbp]
  \centering
  \includegraphics[width=\textwidth]{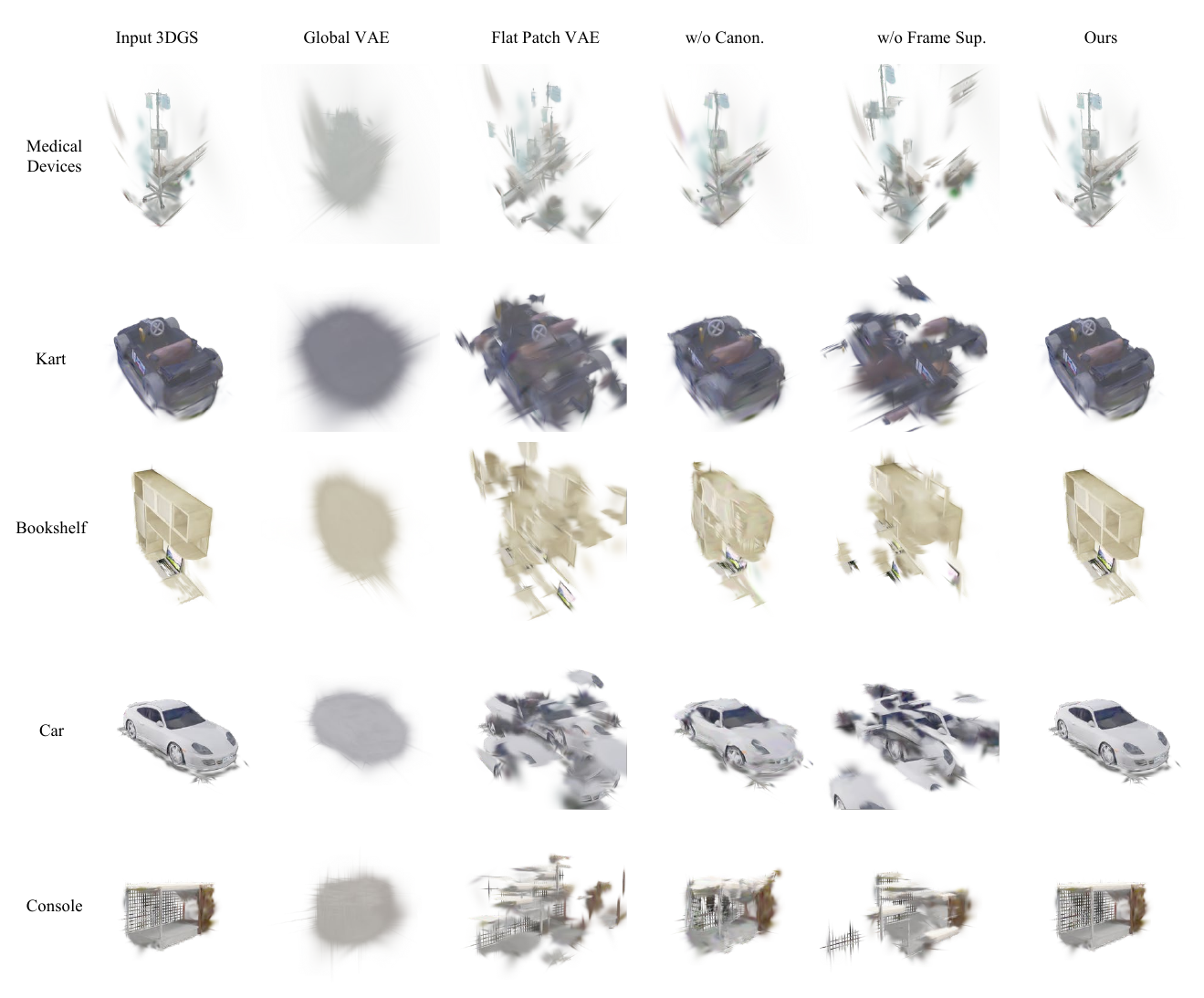}
  \caption{Qualitative comparison of 3DGS reconstruction results using different autoencoding representations. The examples cover structurally different categories, including medical devices, karts, bookshelves, cars, and consoles.}
  \label{fig:vae_reconstruction}
\end{figure}

Fig.~\ref{fig:vae_reconstruction} further visualizes the reconstruction quality of different autoencoding representations, the global VAE produces highly diffuse reconstructions and loses most category-specific geometry, such as the thin supports of the medical device, the shelf partitions, and the car body contour. The flat patch VAE recovers more local appearance cues, but its patch layout is unstable, leading to duplicated fragments and broken object structures. Removing canonicalization weakens the alignment among local patches, so the reconstructed objects preserve the rough silhouette but exhibit distorted fine details and inconsistent local placement. Removing frame supervision causes more severe spatial drift, especially for objects with articulated parts or repeated structures. By comparison, the proposed hierarchical VAE maintains the global object layout while preserving local Gaussian details, resulting in reconstructions that are visually closest to the input 3DGS objects across all tested categories.

\subsection{Comparison with State-of-the-Art Methods}
Table~\ref{tab:main_quantitative} reports the quantitative comparison with representative state-of-the-art methods. The proposed method achieves the best performance across all evaluation aspects, including category consistency, geometric fidelity, rendering quality, multi-view consistency, and inference efficiency. Compared with direct 3DGS adaptations of image-space generative backbones, our method shows a clear advantage in geometry, rendering quality, and efficiency. This is because DiT-3DGS, U-ViT-3DGS, and Vanilla RF-3DGS treat Gaussian patch latents mainly as generic high-dimensional tokens or grids, and therefore do not explicitly encode canonical patch structure, global-local latent coupling, or density-aware rendering importance.

\begin{table}[t]
\caption{Quantitative comparison with state-of-the-art methods. Best results are shown in bold.}
\label{tab:main_quantitative}
\centering
\scriptsize
\setlength{\tabcolsep}{4pt}
\begin{tabular*}{\textwidth}{@{\extracolsep{\fill}}lcccccccccc@{}}
\toprule
\multirow{2}{*}{Method}
& \multicolumn{2}{c}{Semantic}
& \multicolumn{2}{c}{Geometry}
& \multicolumn{2}{c}{Rendering}
& \multicolumn{2}{c}{Consistency}
& \multicolumn{2}{c}{Efficiency} \\
\cmidrule(lr){2-3}
\cmidrule(lr){4-5}
\cmidrule(lr){6-7}
\cmidrule(lr){8-9}
\cmidrule(lr){10-11}
& CAcc $\uparrow$ & CLIP $\uparrow$
& CD $\downarrow$ & F@1 $\uparrow$
& LPIPS $\downarrow$ & SSIM $\uparrow$
& MVC $\uparrow$ & DCS $\uparrow$
& Time $\downarrow$ & Mem. $\downarrow$ \\
\midrule
DiT-3DGS & 83.2 & 0.312 & 3.91 & 57.6 & 0.207 & 0.724 & 0.746 & 0.711 & 35.4 & 10.8 \\
U-ViT-3DGS & 84.5 & 0.319 & 3.72 & 59.1 & 0.198 & 0.737 & 0.759 & 0.728 & 31.7 & 10.1 \\
Vanilla RF-3DGS & 85.8 & 0.327 & 3.56 & 61.0 & 0.189 & 0.751 & 0.774 & 0.742 & 16.9 & 8.9 \\
GaussianCube & 86.1 & 0.329 & 3.42 & 62.7 & 0.183 & 0.758 & 0.791 & 0.754 & 18.6 & 8.4 \\
DiffGS & 87.3 & 0.336 & 3.18 & 64.1 & 0.171 & 0.771 & 0.806 & 0.768 & 24.8 & 9.1 \\
SplatFlow & 88.0 & 0.342 & 3.05 & 65.5 & 0.164 & 0.784 & 0.823 & 0.781 & 11.2 & 7.6 \\
Ours & \textbf{91.6} & \textbf{0.371} & \textbf{2.47} & \textbf{71.8} & \textbf{0.132} & \textbf{0.823} & \textbf{0.872} & \textbf{0.836} & \textbf{3.8} & \textbf{5.2} \\
\bottomrule
\end{tabular*}
\par\vspace{0.3em}
\parbox{0.95\textwidth}{\footnotesize \textit{Note}: CAcc: Category Accuracy (\%); CLIP: CLIP-based category alignment; CD: Chamfer Distance ($10^{-3}$); F@1: F-score at 1\% distance threshold (\%); LPIPS: Learned Perceptual Image Patch Similarity; SSIM: Structural Similarity; MVC: Multi-view Consistency; DCS: Depth Consistency Score; Time: seconds per object; Mem.: GPU memory usage (GB).}
\end{table}

Among the direct image-backbone adaptations, DiT-3DGS and U-ViT-3DGS provide reasonable category alignment but show weaker geometry and multi-view consistency, because their token modeling is inherited from 2D image generation and does not explicitly preserve 3D Gaussian patch correspondence. Vanilla RF-3DGS improves sampling efficiency by replacing diffusion denoising with a simple rectified-flow objective, but its single-level latent modeling still lacks the global-local coupling and rendering-aware weighting required by structured 3DGS objects. GaussianCube and DiffGS directly model Gaussian representations, yet their structured domains are either grid-based or function-based, which can weaken local primitive correspondence and fine patch-level geometry. SplatFlow further improves efficiency by using rectified flow, but its generation process is built on multi-view intermediate latents rather than directly transporting object-level Gaussian patch latents. In contrast, the proposed method performs generation in the hierarchical global-local latent space of 3DGS objects. The lower CD and higher F@1 indicate that the generated objects better preserve 3D structure, while the lower LPIPS and higher SSIM show improved rendered appearance. The higher MVC and DCS further demonstrate stronger cross-view geometry and appearance consistency. Meanwhile, the reduced inference time and memory usage confirm the benefit of structure-aware direct latent-space generation.

\begin{figure}[thbp]
  \centering
  \includegraphics[width=\textwidth]{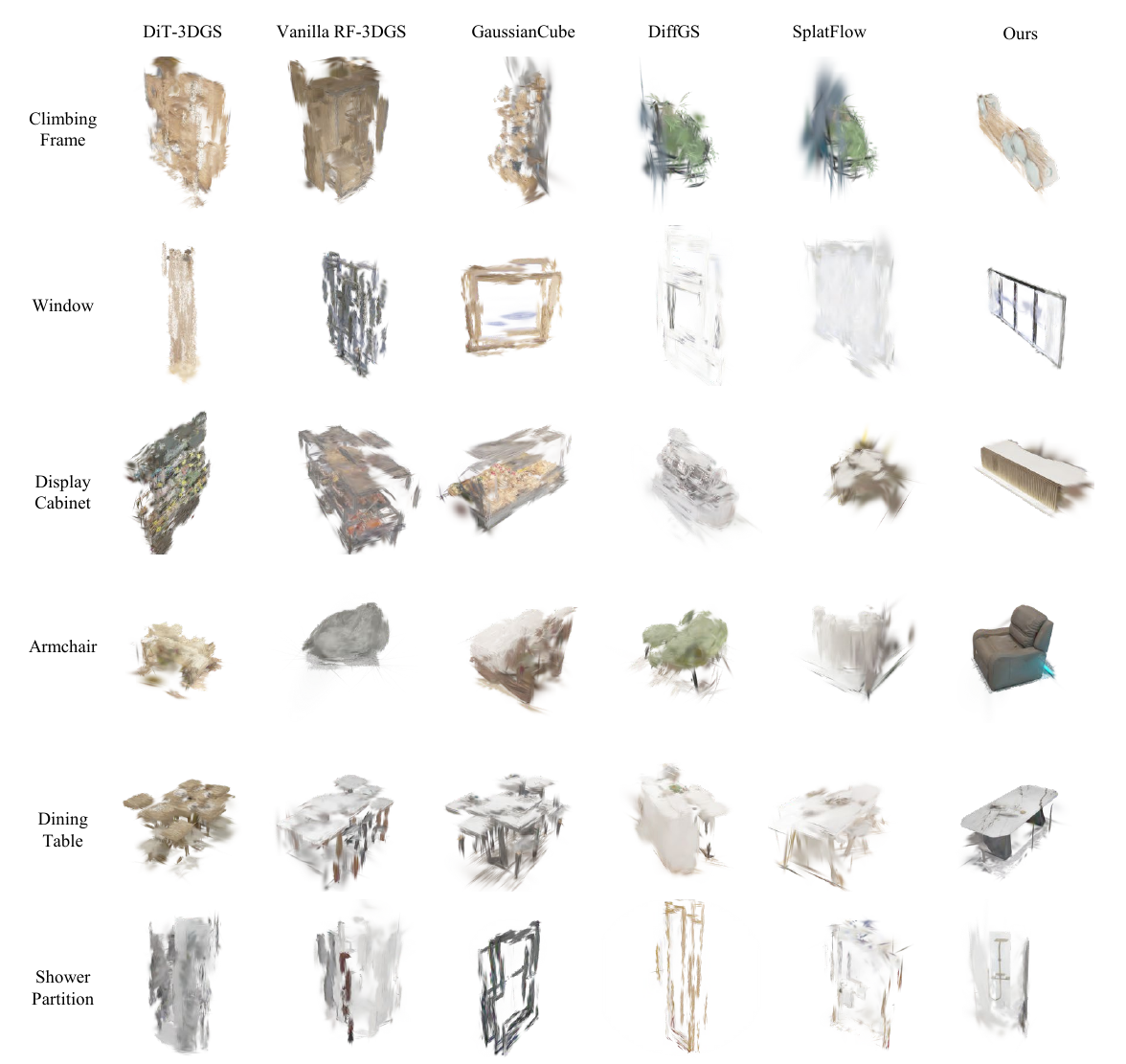}
  \caption{Qualitative comparison with representative 3DGS generation baselines under the same class conditions. The examples include climbing frame, window, display cabinet, armchair, dining table, and shower partition categories.}
  \label{fig:main_qualitative}
\end{figure}

Fig.~\ref{fig:main_qualitative} provides qualitative comparisons under representative class conditions. The direct image-backbone adaptations, such as DiT-3DGS and Vanilla RF-3DGS, can capture coarse category cues but often generate over-smoothed or fragmented Gaussian distributions. This is particularly visible for thin or frame-like objects, where the climbing frame and window results lose clear structural boundaries. GaussianCube and DiffGS produce more recognizable object layouts, but their outputs still contain scattered primitives, incomplete parts, or blurred local geometry, especially for the display cabinet and shower partition categories. SplatFlow improves global plausibility in several cases, yet its multi-view intermediate representation tends to weaken fine 3D correspondence and produces diffuse surfaces for objects with repeated thin structures. In contrast, the proposed method generates more compact and complete 3DGS objects with clearer edges, more stable layouts, and better preserved category-specific details. The armchair, dining table, and window examples show that our hierarchical Gaussian patch latent space better maintains both object-level shape and local geometric structure, which is consistent with the quantitative improvements in CD, F@1, LPIPS, SSIM, and multi-view consistency.

\subsection{Ablation Studies}
\begin{figure}[thbp]
  \centering
  \includegraphics[width=\textwidth]{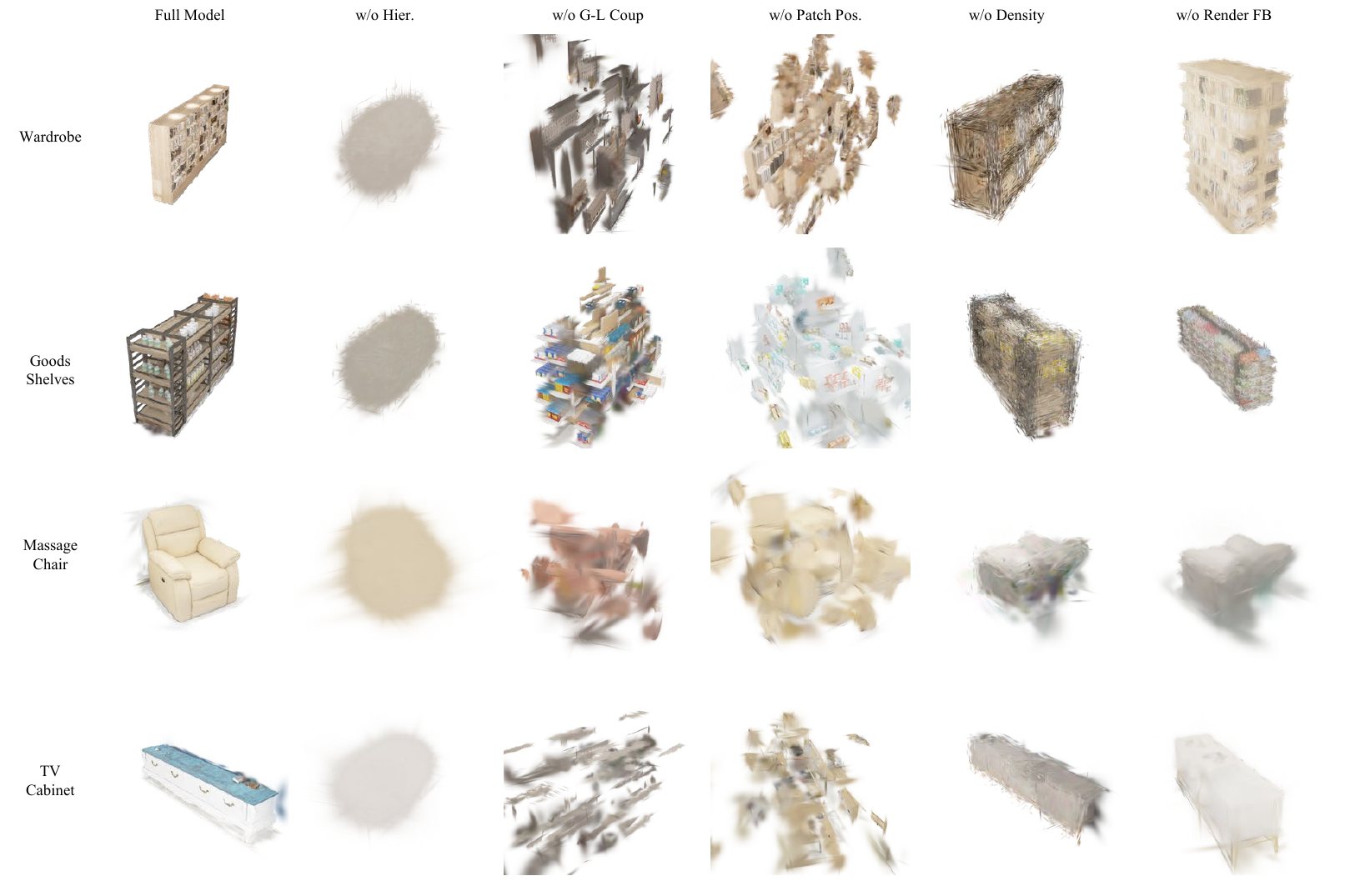}
  \caption{Qualitative visualization of component ablation results. The comparison includes the full model and variants without hierarchical representation, global-local coupling, patch-position conditioning, density-aware weighting, and render-feedback supervision.}
  \label{fig:ablation_visualization}
\end{figure}

We perform ablation studies to validate the effectiveness of each major component, including the hierarchical Gaussian patch representation, structure-aware rectified flow design, and render-feedback supervision. Table~\ref{tab:ablation_main} reports the main component-level ablation results, and Table~\ref{tab:ablation_training} further analyzes training strategies and implementation choices. Fig.~\ref{fig:ablation_visualization} provides qualitative ablation visualization.

\begin{table}[b]
\caption{Main ablation study on the proposed components. Best results are shown in bold.}
\label{tab:ablation_main}
\centering
\scriptsize
\begin{tabular*}{\textwidth}{@{\extracolsep{\fill}}lccccc@{}}
\toprule
Variant & CD $\downarrow$ & LPIPS $\downarrow$ & MVC $\uparrow$ & Time $\downarrow$ & CAcc $\uparrow$ \\
\midrule
Full model & \textbf{2.47} & \textbf{0.132} & \textbf{0.872} & \textbf{3.8} & \textbf{91.6} \\
w/o hierarchical representation & 3.18 & 0.171 & 0.806 & 4.6 & 87.4 \\
w/o global-local coupling & 2.94 & 0.158 & 0.829 & 4.1 & 88.6 \\
w/o patch-position conditioning & 2.83 & 0.153 & 0.817 & 3.9 & 88.1 \\
w/o density-aware weighting & 2.72 & 0.147 & 0.842 & 3.8 & 89.3 \\
w/o render-feedback & 2.68 & 0.151 & 0.833 & 3.7 & 89.0 \\
\bottomrule
\end{tabular*}
\par\vspace{0.3em}
\parbox{0.95\textwidth}{\footnotesize \textit{Note}: CD is reported in $10^{-3}$; Time denotes seconds per generated object.}
\end{table}

\begin{table}[t]
\caption{Additional ablation study on training strategy and implementation choices. Best results are shown in bold.}
\label{tab:ablation_training}
\centering
\scriptsize
\begin{tabular*}{\textwidth}{@{\extracolsep{\fill}}lccccc@{}}
\toprule
Setting & CD $\downarrow$ & LPIPS $\downarrow$ & MVC $\uparrow$ & Stability $\uparrow$ & CAcc $\uparrow$ \\
\midrule
Two-stage training & \textbf{2.47} & \textbf{0.132} & \textbf{0.872} & \textbf{0.94} & \textbf{91.6} \\
Joint training & 2.91 & 0.158 & 0.824 & 0.78 & 88.2 \\
Frozen decoder in stage 2 & 2.69 & 0.146 & 0.846 & 0.90 & 89.4 \\
$M=128$ patches & 2.78 & 0.149 & 0.839 & 0.88 & 89.1 \\
$M=512$ patches & 2.55 & 0.137 & 0.861 & 0.86 & 90.4 \\
$S=16$ ODE steps & 2.63 & 0.142 & 0.853 & 0.93 & 90.1 \\
$S=64$ ODE steps & 2.48 & 0.133 & 0.871 & 0.91 & 91.5 \\
\bottomrule
\end{tabular*}
\par\vspace{0.3em}
\parbox{0.95\textwidth}{\footnotesize \textit{Note}: Stability is a normalized training-stability score measured from loss variance and convergence success rate; higher is better.}
\end{table}

Table~\ref{tab:ablation_main} shows that each proposed component contributes to the final generation quality. Removing the hierarchical representation causes the largest performance drop, increasing CD from 2.47 to 3.18 and LPIPS from 0.132 to 0.171, while reducing MVC and CAcc to 0.806 and 87.4, respectively. This confirms that a single-level or structure-agnostic latent space is insufficient for preserving both object-level layout and local Gaussian details. Removing global-local coupling also leads to a clear degradation, indicating that global semantic layout and local patch generation should not be modeled independently. Without patch-position conditioning, the model obtains lower MVC than the variant without global-local coupling, which suggests that explicit spatial anchors are especially important for preserving cross-view consistency and repeated structures. Density-aware weighting mainly affects rendering quality and geometric fidelity, because visually important or high-contribution Gaussian primitives are no longer emphasized during flow learning. Removing render-feedback also increases LPIPS and decreases MVC, showing that latent-space distribution matching alone cannot fully guarantee view-consistent rendered appearance after decoding.

Table~\ref{tab:ablation_training} further analyzes training and implementation choices. The two-stage training strategy achieves the best overall balance, with the lowest CD and LPIPS, the highest MVC, and the most stable convergence. Joint training performs noticeably worse and has the lowest stability score, suggesting that simultaneously learning the VAE reconstruction space and the generative flow can cause optimization interference. Freezing the decoder in the second stage improves stability compared with joint training, but it limits the decoder's ability to adapt to generated latent samples, resulting in worse rendering and consistency than the full two-stage setting. The patch-number study shows that using too few patches ($M=128$) under-represents fine local structures, whereas increasing the number of patches can further improve geometric quality at the cost of higher computation. We choose $M=256$ as the default because it achieves performance close to larger patch configurations while maintaining substantially better inference efficiency and training stability. For ODE sampling, reducing the integration steps to $S=16$ improves neither quality nor consistency, while increasing to $S=64$ provides only marginal changes at higher sampling cost. Therefore, the default configuration offers a better trade-off between reconstruction fidelity, rendering quality, multi-view consistency, and training stability.

Fig.~\ref{fig:ablation_visualization} further illustrates the role of each component. The full model generates compact and recognizable Gaussian objects with clear category-specific structures, such as the layered wardrobe, goods shelves, massage chair, and TV cabinet. Without the hierarchical representation, the generated objects collapse into diffuse Gaussian blobs, which visually confirms the severe degradation observed in Table~\ref{tab:ablation_main}. Removing global-local coupling preserves some local fragments but breaks the object-level organization, leading to misplaced shelves, distorted chair parts, and unstable cabinet layouts. Without patch-position conditioning, patch locations become less constrained, causing duplicated or drifting structures and weaker spatial coherence. The variant without density-aware weighting keeps the rough object extent but introduces noisy and uneven Gaussian distributions, especially around thin supports, edges, and shelf-like repeated structures. Removing render-feedback produces smoother but less detailed objects, indicating that latent-space supervision alone is insufficient for preserving rendered appearance. These qualitative results are consistent with the quantitative trends and show that the proposed components jointly improve structural completeness, local detail preservation, and rendering consistency.

\subsection{Qualitative Results}
\begin{figure}[thbp]
  \centering
  \includegraphics[width=\textwidth]{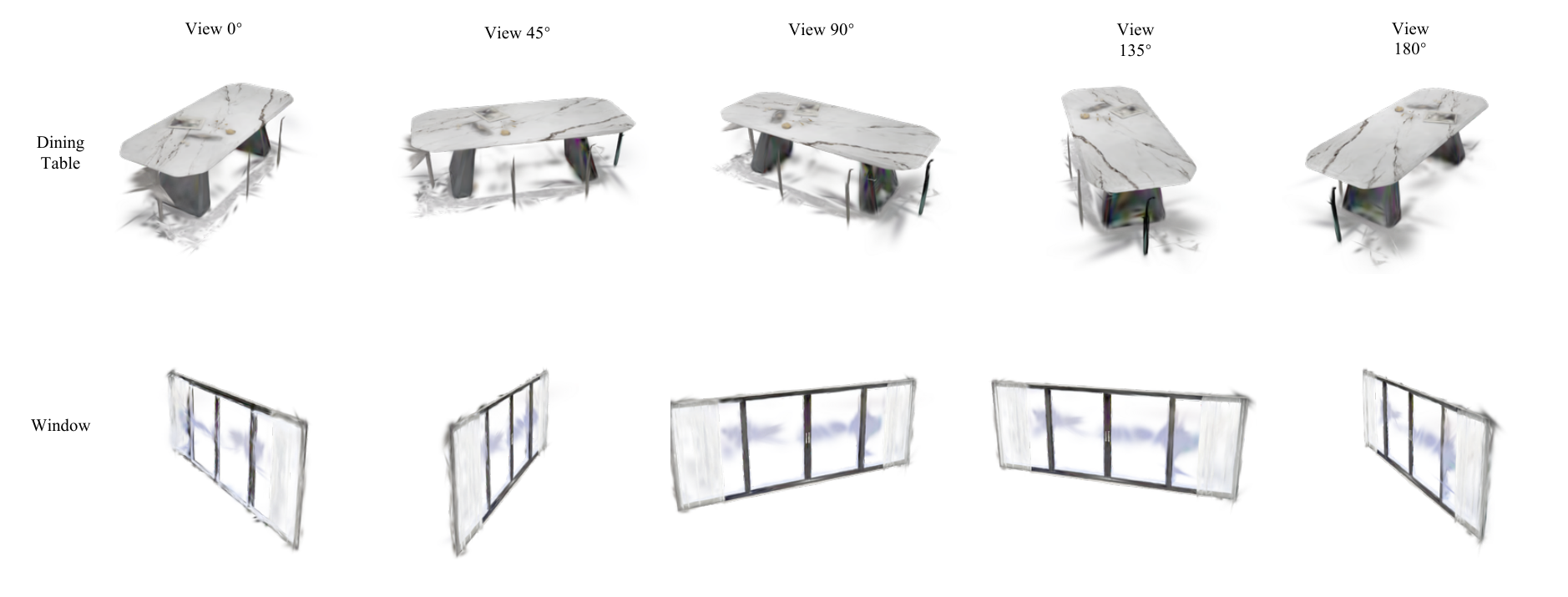}
  \caption{Multi-view rendering results of generated 3D Gaussian objects. Each row shows the same generated object rendered from five viewpoints, including $0^\circ$, $45^\circ$, $90^\circ$, $135^\circ$, and $180^\circ$.}
  \label{fig:multiview_results}
\end{figure}

\begin{figure}[thbp]
  \centering
  \includegraphics[width=\textwidth]{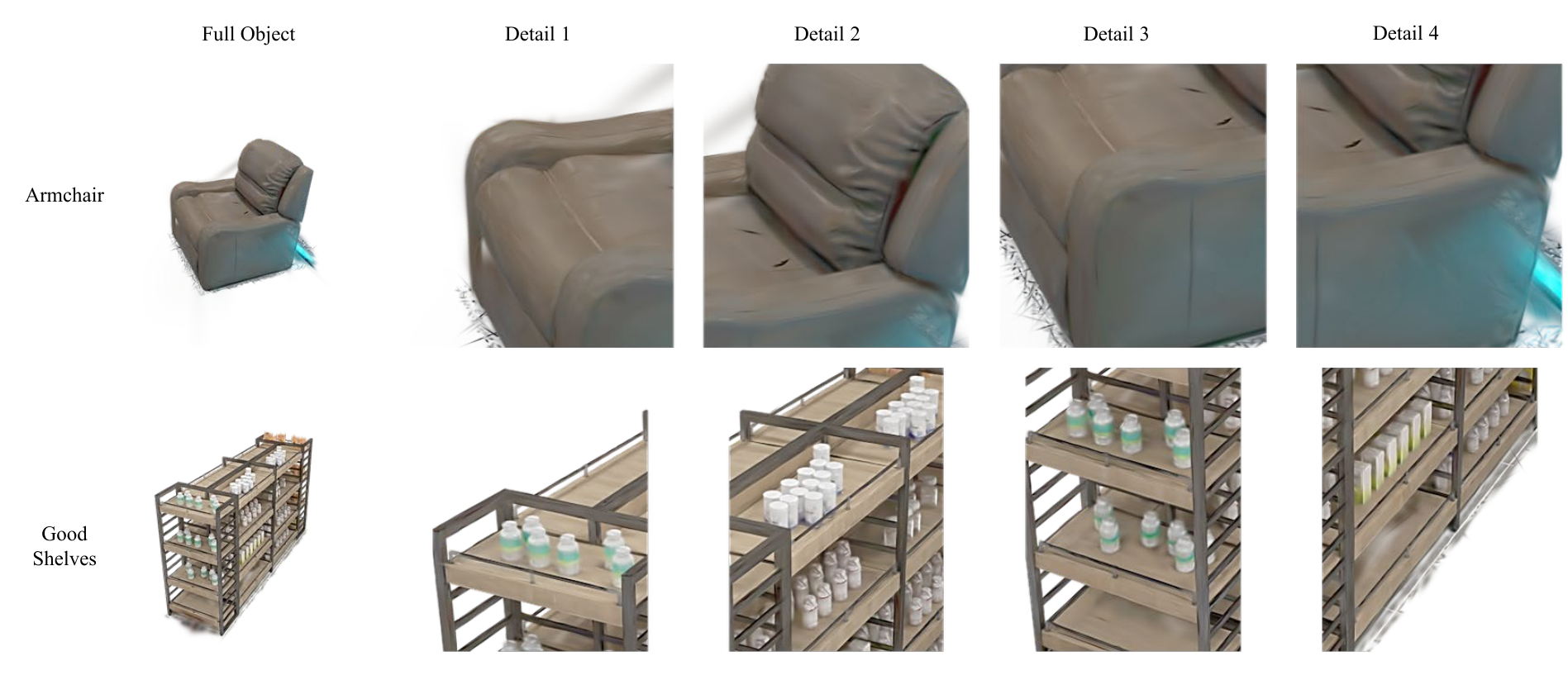}
  \caption{Local detail visualization of generated 3D Gaussian objects. Each row shows a full generated object and four zoomed-in regions that contain texture-sensitive surfaces, object boundaries, repeated small parts, or thin structural elements.}
  \label{fig:local_detail_results}
\end{figure}

This section presents additional visual results of generated 3D Gaussian objects, including multi-view rendering and local detail visualization. Fig.~\ref{fig:multiview_results} shows multi-view rendering results, and Fig.~\ref{fig:local_detail_results} is reserved for local detail close-up visualization.

Fig.~\ref{fig:multiview_results} demonstrates that the generated 3DGS objects maintain coherent geometry and appearance under large viewpoint changes. For the dining table, the tabletop shape, supporting legs, and material pattern remain consistent from oblique, side, and frontal views, indicating that the model learns an object-level 3D structure rather than a view-specific appearance. For the window example, the thin frame bars and transparent panel regions are preserved across viewpoints, which is challenging because small misalignments in Gaussian positions can easily cause broken or drifting line structures. These results provide visual evidence for the improved multi-view consistency reported in Table~\ref{tab:main_quantitative}, and further show that the hierarchical Gaussian patch latent space can generate objects with stable spatial layouts and renderable local details.

Fig.~\ref{fig:local_detail_results} presents local close-up visualizations for the armchair and goods shelves categories. In the armchair example, the generated object preserves smooth cushion surfaces, clear seam-like boundaries, and coherent transitions between the seat, armrest, and backrest. These details indicate that the decoder can recover local geometric variation from the hierarchical Gaussian patch latents rather than producing only a coarse object silhouette. In the goods shelves example, the shelf frames, layered boards, and repeated bottle-like objects remain distinguishable in the enlarged regions. This is important because repeated thin structures and small objects are sensitive to Gaussian position drift and density imbalance. The close-up results show that the proposed method can maintain compact local Gaussian distributions and preserve fine rendering details, complementing the quantitative improvements in LPIPS, SSIM, CD, and F@1.

\subsection{Further Analysis}
\begin{figure}[thbp]
  \centering
  \includegraphics[width=\textwidth]{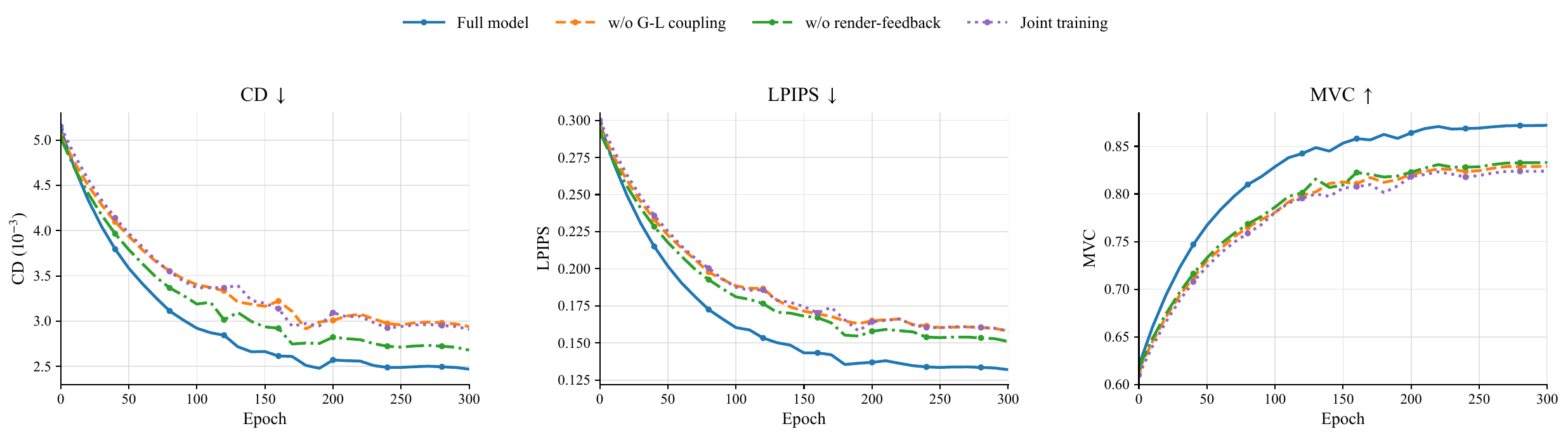}
  \caption{Training convergence curves of the proposed method and representative variants. CD, LPIPS, and MVC are evaluated during training to compare convergence speed, final generation quality, and multi-view consistency.}
  \label{fig:convergence_curve}
\end{figure}

\begin{figure}[thbp]
  \centering
  \includegraphics[width=\textwidth]{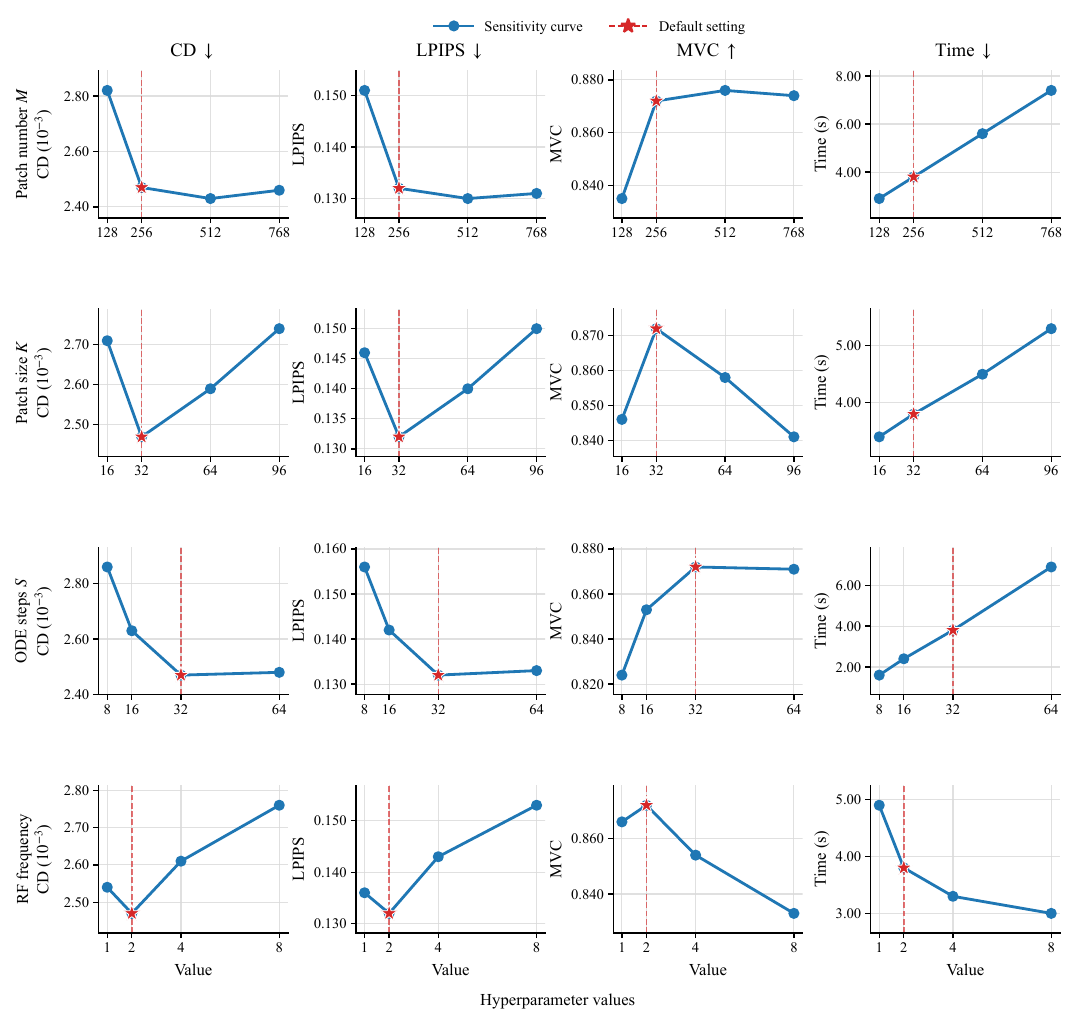}
  \caption{Sensitivity analysis of key hyperparameters, including patch number $M$, patch size $K$, ODE step number $S$, and render-feedback frequency. The red star denotes the default setting used in the main experiments.}
  \label{fig:sensitivity_analysis}
\end{figure}

We further analyze the convergence behavior and hyperparameter sensitivity of the proposed method. Fig.~\ref{fig:convergence_curve} reports training convergence curves, and Fig.~\ref{fig:sensitivity_analysis} presents hyperparameter sensitivity analysis.

Fig.~\ref{fig:convergence_curve} shows that the full model converges faster and reaches better final performance than the representative variants. For CD and LPIPS, the full model decreases more rapidly during the early training stage and remains comparatively stable despite moderate mid-training oscillations, indicating that the hierarchical representation and render-feedback supervision provide a more reliable optimization target. The variants without global-local coupling or render-feedback exhibit larger fluctuations and converge to higher CD and LPIPS values, showing that removing either structural coupling or rendering-aware supervision weakens the learned latent distribution. For MVC, the full model consistently achieves higher multi-view consistency throughout training, while joint training and the ablated variants fluctuate more noticeably before saturating at lower values. These results support the two-stage training design and confirm that the proposed components not only improve final quality but also make the training process more stable and effective.

Fig.~\ref{fig:sensitivity_analysis} analyzes how key hyperparameters affect geometric fidelity, rendering quality, multi-view consistency, and inference time. Increasing the patch number $M$ improves the representation capacity and leads to better CD, LPIPS, and MVC up to a moderate range. However, larger values such as $M=512$ and $M=768$ introduce substantially higher inference cost, while the quality gain over $M=256$ is relatively small. Therefore, $M=256$ is selected as the default setting because it provides a favorable quality-efficiency trade-off. For the patch size $K$, the best results are obtained around $K=32$; smaller patches lack sufficient local context, whereas overly large patches weaken locality and increase computation. For the ODE step number $S$, increasing the number of steps improves sampling quality from $S=8$ to $S=32$, but further increasing to $S=64$ brings negligible quality gains with much higher inference time. The render-feedback frequency also shows a clear trade-off: applying feedback too frequently increases cost, while sparse feedback weakens rendering supervision. Applying render-feedback every two flow iterations gives the best balance between rendering quality, multi-view consistency, and efficiency.

\section{Discussion}

\textbf{Structured Gaussian patch representation.} The experimental results indicate that structured Gaussian patch modeling is essential for direct 3DGS object generation. Unlike point clouds or dense grids, a 3D Gaussian object contains irregular primitives with unequal rendering contributions, local density variation, and strong dependence between spatial layout and appearance attributes. A purely global latent representation can capture coarse category semantics, but it tends to lose local geometric details and thin structures. In contrast, the proposed hierarchical Gaussian patch representation decomposes the object into canonical local patches while preserving a global semantic latent code. This design allows the model to jointly represent object-level layout and patch-level geometry, which explains the consistent improvements in reconstruction quality, generation fidelity, and multi-view consistency.

\textbf{Direct latent generation of 3DGS objects.} Another important observation is that direct latent generation of 3D Gaussian objects differs substantially from lifting image-space generative models to 3DGS. Image diffusion backbones and transformer-based token generators are effective in 2D domains, but their token structures do not naturally encode 3D Gaussian patch correspondence, density distribution, or rendering-sensitive primitive attributes. When such models are directly adapted to 3DGS latents, they can learn coarse category patterns but often produce fragmented structures or unstable cross-view geometry. The proposed method instead learns the intrinsic class-conditioned distribution of structured 3DGS objects, avoiding per-instance optimization and reducing dependence on multi-view intermediate representations. This makes the framework more suitable for efficient class-guided object synthesis.

\textbf{Render-aware latent flow learning.} The render-feedback strategy further bridges the gap between latent-space generation and final rendered quality. Since the same latent error can produce different visual effects after Gaussian decoding and rendering, optimizing only a latent velocity loss may not guarantee perceptually coherent multi-view outputs. By introducing rendering-aware supervision during flow learning, the generated latents are encouraged to decode into objects that are not only close to the latent distribution but also visually consistent under different viewpoints. The ablation results show that this feedback is especially useful for preserving local details, reducing view-dependent artifacts, and stabilizing thin or repeated structures.

\textbf{Quality-efficiency trade-off.} The sensitivity analysis also reveals a practical quality-efficiency trade-off. Increasing the patch number improves representational capacity and can slightly improve geometric and rendering metrics, but the benefit becomes marginal once the patch structure is sufficiently expressive. Larger patch numbers also increase memory usage and inference cost. Therefore, the default setting of $M=256$ is chosen because it achieves performance close to larger configurations while maintaining better efficiency. A similar trade-off is observed for the number of ODE steps and render-feedback frequency: more computation can improve generation quality to some extent, but excessive sampling steps or overly frequent rendering feedback provide limited additional gains. These results suggest that the proposed framework benefits from structured latent modeling rather than relying mainly on heavy sampling or rendering procedures.

\textbf{Limitations and future work.} Despite these advantages, several limitations remain. First, the current framework focuses on class-guided generation and does not yet support open-vocabulary text-conditioned 3DGS synthesis. Incorporating language embeddings or multimodal priors could extend the method to more flexible generation scenarios. Second, the method relies on object-level 3DGS preprocessing and assumes that each object can be normalized and decomposed into a shared patch-anchor structure. This setting is effective for object generation, but complex scene-level generation with strong object interactions may require adaptive patch allocation or hierarchical scene decomposition. Third, highly thin, transparent, reflective, or topologically complex objects may still suffer from local Gaussian drift or incomplete fine structures. Future work could explore variable-size patch representations, stronger physical or material-aware constraints, and joint modeling of object-level and scene-level Gaussian distributions.

\section{Conclusion}

This paper presented a hierarchical Gaussian patch representation for direct class-guided 3D Gaussian object generation. By decomposing irregular Gaussian primitives into canonical local patches and organizing them with a global-local latent structure, the proposed method provides a structured representation space that preserves both object-level semantics and local rendering-sensitive details. Based on this representation, we developed a class-guided structure-aware rectified flow model with patch-position conditioning, global-local coupled velocity prediction, and density-aware velocity weighting. A render-feedback learning strategy was further introduced to connect latent-space flow optimization with decoded multi-view rendering quality.

Extensive experiments demonstrate that the proposed framework generates more coherent, detailed, and view-consistent 3D Gaussian objects than representative 3DGS generation baselines and direct adaptations of image-space generative backbones. The ablation studies confirm the effectiveness of the hierarchical representation, global-local coupling, patch-position conditioning, density-aware weighting, render-feedback supervision, and two-stage training strategy. The convergence and sensitivity analyses further show that the method achieves a favorable balance between generation quality and computational efficiency. These results suggest that explicitly modeling the intrinsic structured distribution of 3DGS objects is a promising direction for efficient 3D generative modeling.








%


\printbibliography

@article{li2026mvgsplatting,
  title={{MVG-Splatting}: Multi-View Guided Gaussian Splatting with Adaptive Quantile-Based Geometric Consistency Densification},
  author={Li, Zhuoxiao and Yao, Shanliang and Chu, Yijie and Garc{\'i}a-Fern{\'a}ndez, {\'A}ngel F. and Yue, Yong and Ding, Weiping and Zhu, Xiaohui},
  journal={Information Fusion},
  volume={126},
  pages={103540},
  year={2026},
  doi={10.1016/j.inffus.2025.103540}
}

@article{ning2024dilf,
  title={{DILF}: Differentiable Rendering-Based Multi-View Image--Language Fusion for Zero-Shot 3D Shape Understanding},
  author={Ning, Xin and Yu, Zaiyang and Li, Lusi and Li, Weijun and Tiwari, Prayag},
  journal={Information Fusion},
  volume={102},
  pages={102033},
  year={2024},
  doi={10.1016/j.inffus.2023.102033}
}

@article{hao2024coarse,
  title={Coarse to Fine-Based Image--Point Cloud Fusion Network for 3D Object Detection},
  author={Hao, Meilan and Zhang, Zhongkang and Li, Lei and Dong, Kejian and Cheng, Long and Tiwari, Prayag and Ning, Xin},
  journal={Information Fusion},
  volume={112},
  pages={102551},
  year={2024},
  doi={10.1016/j.inffus.2024.102551}
}

@article{sohail2025advancing,
  title={Advancing 3D Point Cloud Understanding Through Deep Transfer Learning: A Comprehensive Survey},
  author={Sohail, Shahab Saquib and Himeur, Yassine and Kheddar, Hamza and Amira, Abbes and Fadli, Fodil and Atalla, Shadi and Copiaco, Abigail and Mansoor, Wathiq},
  journal={Information Fusion},
  volume={113},
  pages={102601},
  year={2025},
  doi={10.1016/j.inffus.2024.102601}
}

@inproceedings{mildenhall2020nerf,
  title={NeRF: Representing Scenes as Neural Radiance Fields for View Synthesis},
  author={Mildenhall, Ben and Srinivasan, Pratul P. and Tancik, Matthew and Barron, Jonathan T. and Ramamoorthi, Ravi and Ng, Ren},
  booktitle={European Conference on Computer Vision},
  pages={405--421},
  year={2020}
}

@article{kerbl2023gaussian,
  title={3D Gaussian Splatting for Real-Time Radiance Field Rendering},
  author={Kerbl, Bernhard and Kopanas, Georgios and Leimk{\"u}hler, Thomas and Drettakis, George},
  journal={ACM Transactions on Graphics},
  volume={42},
  number={4},
  pages={139:1--139:14},
  year={2023}
}

@inproceedings{ho2020denoising,
  title={Denoising Diffusion Probabilistic Models},
  author={Ho, Jonathan and Jain, Ajay and Abbeel, Pieter},
  booktitle={Advances in Neural Information Processing Systems},
  volume={33},
  pages={6840--6851},
  year={2020}
}

@inproceedings{rombach2022high,
  title={High-Resolution Image Synthesis with Latent Diffusion Models},
  author={Rombach, Robin and Blattmann, Andreas and Lorenz, Dominik and Esser, Patrick and Ommer, Bj{\"o}rn},
  booktitle={Proceedings of the IEEE/CVF Conference on Computer Vision and Pattern Recognition},
  pages={10684--10695},
  year={2022}
}

@inproceedings{poole2023dreamfusion,
  title={DreamFusion: Text-to-3D using 2D Diffusion},
  author={Poole, Ben and Jain, Ajay and Barron, Jonathan T. and Mildenhall, Ben},
  booktitle={International Conference on Learning Representations},
  year={2023}
}

@inproceedings{shi2024mvdream,
  title={MVDream: Multi-view Diffusion for 3D Generation},
  author={Shi, Yichun and Wang, Peng and Ye, Jianglong and Long, Mai and Li, Kejie and Yang, Xiao},
  booktitle={International Conference on Learning Representations},
  year={2024}
}

@article{liu2025maenet,
  title={{MAENet}: Boost Image-Guided Point Cloud Completion More Accurate and Even},
  author={Liu, Moyun and Yang, Ziheng and Chen, Bing and Chen, Youping and Xie, Jingming and Yao, Lei and Chau, Lap-Pui and Du, Jiawei and Zhou, Joey Tianyi},
  journal={Information Fusion},
  volume={122},
  pages={103179},
  year={2025},
  doi={10.1016/j.inffus.2025.103179}
}

@article{wu2026fusionmamba,
  title={FusionMamba: Efficient 3D Point Cloud Completion via Spatially-Aware State Space Models and Multi-Scale Information Fusion},
  author={Wu, Xiang and Wang, Junchao and Zhang, Shijie and Chen, Tianfei and Liu, Xiaoping and He, Shuyao},
  journal={Information Fusion},
  volume={134},
  pages={104378},
  year={2026},
  doi={10.1016/j.inffus.2026.104378}
}

@inproceedings{tang2024dreamgaussian,
  title={DreamGaussian: Generative Gaussian Splatting for Efficient 3D Content Creation},
  author={Tang, Jiaxiang and Ren, Jiawei and Zhou, Hang and Liu, Ziwei and Zeng, Gang},
  booktitle={International Conference on Learning Representations},
  year={2024}
}

@inproceedings{chen2024text3d,
  title={Text-to-3D using Gaussian Splatting},
  author={Chen, Zilong and Wang, Feng and Wang, Yikai and Liu, Huaping},
  booktitle={Proceedings of the IEEE/CVF Conference on Computer Vision and Pattern Recognition},
  pages={21401--21412},
  year={2024}
}

@inproceedings{yi2024gaussiandreamer,
  title={GaussianDreamer: Fast Generation from Text to 3D Gaussians by Bridging 2D and 3D Diffusion Models},
  author={Yi, Taoran and Fang, Jiemin and Wang, Junjie and Wu, Guanjun and Xie, Lingxi and Zhang, Xiaopeng and Liu, Wenyu and Tian, Qi and Wang, Xinggang},
  booktitle={Proceedings of the IEEE/CVF Conference on Computer Vision and Pattern Recognition},
  pages={6796--6807},
  year={2024}
}

@inproceedings{szymanowicz2024splatter,
  title={Splatter Image: Ultra-Fast Single-View 3D Reconstruction},
  author={Szymanowicz, Stanislaw and Rupprecht, Christian and Vedaldi, Andrea},
  booktitle={Proceedings of the IEEE/CVF Conference on Computer Vision and Pattern Recognition},
  year={2024}
}

@inproceedings{zou2024triplanegaussian,
  title={Triplane Meets Gaussian Splatting: Fast and Generalizable Single-View 3D Reconstruction with Transformers},
  author={Zou, Zi-Xin and Yu, Zhipeng and Guo, Yuan-Chen and Li, Yangguang and Liang, Ding and Cao, Yan-Pei and Zhang, Song-Hai},
  booktitle={Proceedings of the IEEE/CVF Conference on Computer Vision and Pattern Recognition},
  pages={10324--10335},
  year={2024}
}

@inproceedings{tang2024lgm,
  title={LGM: Large Multi-View Gaussian Model for High-Resolution 3D Content Creation},
  author={Tang, Jiaxiang and Chen, Zhaoxi and Chen, Xiaokang and Wang, Tengfei and Zeng, Gang and Liu, Ziwei},
  booktitle={European Conference on Computer Vision},
  year={2024}
}

@inproceedings{xu2024grm,
  title={GRM: Large Gaussian Reconstruction Model for Efficient 3D Reconstruction and Generation},
  author={Xu, Yinghao and Shi, Zifan and Wang, Yifan and Chen, Hansheng and Yang, Ceyuan and Peng, Sida and Shen, Yujun and Wetzstein, Gordon},
  booktitle={European Conference on Computer Vision},
  year={2024}
}

@inproceedings{zhang2024gaussiancube,
  title={GaussianCube: A Structured and Explicit Radiance Representation for 3D Generative Modeling},
  author={Zhang, Bowen and Cheng, Yiji and Yang, Jiaolong and Wang, Chunyu and Zhao, Feng and Tang, Yansong and Chen, Dong and Guo, Baining},
  booktitle={Advances in Neural Information Processing Systems},
  year={2024}
}

@inproceedings{zhou2024diffgs,
  title={DiffGS: Functional Gaussian Splatting Diffusion},
  author={Zhou, Junsheng and Zhang, Weiqi and Liu, Yu-Shen},
  booktitle={Advances in Neural Information Processing Systems},
  year={2024}
}

@inproceedings{lipman2023flowmatching,
  title={Flow Matching for Generative Modeling},
  author={Lipman, Yaron and Chen, Ricky T. Q. and Ben-Hamu, Heli and Nickel, Maximilian and Le, Matt},
  booktitle={International Conference on Learning Representations},
  year={2023}
}

@inproceedings{liu2023flowstraight,
  title={Flow Straight and Fast: Learning to Generate and Transfer Data with Rectified Flow},
  author={Liu, Xingchao and Gong, Chengyue and Liu, Qiang},
  booktitle={International Conference on Learning Representations},
  year={2023}
}

@inproceedings{ju2025directtrigs,
  title={DirectTriGS: Triplane-based Gaussian Splatting Field Representation for 3D Generation},
  author={Ju, Xiaoliang and Li, Hongsheng},
  booktitle={Proceedings of the IEEE/CVF Conference on Computer Vision and Pattern Recognition},
  year={2025}
}

@inproceedings{go2025splatflow,
  title={SplatFlow: Multi-View Rectified Flow Model for 3D Gaussian Splatting Synthesis},
  author={Go, Hyojun and Park, Byeongjun and Jang, Jiho and Kim, Jin-Young and Kwon, Soonwoo and Kim, Changick},
  booktitle={Proceedings of the IEEE/CVF Conference on Computer Vision and Pattern Recognition},
  year={2025}
}

@article{mariani2026glue3d,
  title={{GLUE3D}: General Language Understanding Evaluation for 3D Point Clouds},
  author={Mariani, Giorgio and Raganato, Alessandro and Melzi, Simone and Pasi, Gabriella},
  journal={Information Fusion},
  volume={129},
  pages={104007},
  year={2026},
  doi={10.1016/j.inffus.2025.104007}
}

@article{fernandes2021pointcloud,
  title={Point-Cloud Based 3D Object Detection and Classification Methods for Self-Driving Applications: A Survey and Taxonomy},
  author={Fernandes, Duarte and Silva, Ant{\'o}nio and N{\'e}voa, Rafael and Sim{\~o}es, Cl{\'a}udia and Gonzalez, Dibet and Guevara, Miguel and Novais, Paulo and Monteiro, Jo{\~a}o and Melo-Pinto, Pedro},
  journal={Information Fusion},
  volume={68},
  pages={161--191},
  year={2021},
  doi={10.1016/j.inffus.2020.11.002}
}

@article{zhou2025waterhenerf,
  title={WaterHE-NeRF: Water-Ray Matching Neural Radiance Fields for Underwater Scene Reconstruction},
  author={Zhou, Jingchun and Liang, Tianyu and Zhang, Dehuan and Liu, Siyuan and Wang, Junsheng and Wu, Edmond Q.},
  journal={Information Fusion},
  volume={115},
  pages={102770},
  year={2025},
  doi={10.1016/j.inffus.2024.102770}
}

@inproceedings{charatan2024pixelsplat,
  title={pixelSplat: 3D Gaussian Splats from Image Pairs for Scalable Generalizable 3D Reconstruction},
  author={Charatan, David and Li, Sizhe and Tagliasacchi, Andrea and Sitzmann, Vincent},
  booktitle={Proceedings of the IEEE/CVF Conference on Computer Vision and Pattern Recognition},
  year={2024}
}

@inproceedings{chen2024mvsplat,
  title={MVSplat: Efficient 3D Gaussian Splatting from Sparse Multi-View Images},
  author={Chen, Yuedong and Xu, Haofei and Zheng, Chuanxia and Zhuang, Bohan and Pollefeys, Marc and Geiger, Andreas and Cham, Tat-Jen and Cai, Jianfei},
  booktitle={European Conference on Computer Vision},
  year={2024}
}

@inproceedings{zhang2024gslrm,
  title={GS-LRM: Large Reconstruction Model for 3D Gaussian Splatting},
  author={Zhang, Kai and Bi, Sai and Tan, Hao and Xiangli, Yuanbo and Zhao, Nanxuan and Sunkavalli, Kalyan and Xu, Zexiang},
  booktitle={European Conference on Computer Vision},
  year={2024}
}

@inproceedings{liang2024luciddreamer,
  title={LucidDreamer: Towards High-Fidelity Text-to-3D Generation via Interval Score Matching},
  author={Liang, Yixun and Yang, Xin and Lin, Jiantao and Li, Haodong and Xu, Xiaogang and Chen, Yingcong},
  booktitle={Proceedings of the IEEE/CVF Conference on Computer Vision and Pattern Recognition},
  year={2024}
}

@article{tong2024improving,
  title={Improving and Generalizing Flow-Based Generative Models with Minibatch Optimal Transport},
  author={Tong, Alexander and Fatras, Kilian and Malkin, Nikolay and Huguet, Guillaume and Zhang, Yanlei and Rector-Brooks, Jarrid and Wolf, Guy and Bengio, Yoshua},
  journal={Transactions on Machine Learning Research},
  year={2024}
}

@inproceedings{tong2024simulation,
  title={Simulation-Free Schr{\"o}dinger Bridges via Score and Flow Matching},
  author={Tong, Alexander and Malkin, Nikolay and Fatras, Kilian and Atanackovic, Lazar and Zhang, Yanlei and Huguet, Guillaume and Wolf, Guy and Bengio, Yoshua},
  booktitle={Proceedings of The 27th International Conference on Artificial Intelligence and Statistics},
  pages={1279--1287},
  year={2024}
}

@inproceedings{chen2024rfm,
  title={Flow Matching on General Geometries},
  author={Chen, Ricky T. Q. and Lipman, Yaron},
  booktitle={International Conference on Learning Representations},
  year={2024}
}

@article{chang2015shapenet,
  title={ShapeNet: An Information-Rich 3D Model Repository},
  author={Chang, Angel X. and Funkhouser, Thomas and Guibas, Leonidas and Hanrahan, Pat and Huang, Qixing and Li, Zimo and Savarese, Silvio and Savva, Manolis and Song, Shuran and Su, Hao and Xiao, Jianxiong and Yi, Li and Yu, Fisher},
  journal={arXiv preprint arXiv:1512.03012},
  year={2015}
}

@inproceedings{deitke2023objaverse,
  title={Objaverse: A Universe of Annotated 3D Objects},
  author={Deitke, Matt and Schwenk, Dustin and Salvador, Jordi and Weihs, Luca and Michel, Oscar and VanderBilt, Eli and Schmidt, Ludwig and Ehsani, Kiana and Kembhavi, Aniruddha and Farhadi, Ali},
  booktitle={Proceedings of the IEEE/CVF Conference on Computer Vision and Pattern Recognition},
  pages={13142--13153},
  year={2023}
}

@inproceedings{collins2022abo,
  title={{ABO}: Dataset and Benchmarks for Real-World 3D Object Understanding},
  author={Collins, Jasmine and Goel, Shubham and Deng, Kenan and Luthra, Achleshwar and Xu, Leon and Gundogdu, Erhan and Zhang, Xi and Vicente, Tomas F. Yago and Dideriksen, Thomas and Arora, Himanshu and Guillaumin, Matthieu and Malik, Jitendra},
  booktitle={Proceedings of the IEEE/CVF Conference on Computer Vision and Pattern Recognition},
  pages={21126--21136},
  year={2022}
}

@article{fu20213dfuture,
  title={{3D-FUTURE}: 3D Furniture Shape with Texture},
  author={Fu, Huan and Cai, Bowen and Gao, Lin and Zhang, Lingxiao and Li, Cao and Zeng, Qixun and Sun, Chengyue and Jia, Rongfei and Zhao, Binqiang and Zhang, Hao},
  journal={International Journal of Computer Vision},
  volume={129},
  number={12},
  pages={3313--3337},
  year={2021}
}

@misc{interiorgs2025,
  title={InteriorGS: A 3D Gaussian Splatting Dataset of Semantically Labeled Indoor Scenes},
  author={{SpatialVerse Research Team, Manycore Tech Inc.}},
  year={2025},
  howpublished={\url{https://huggingface.co/datasets/spatialverse/InteriorGS}}
}

\end{document}